\documentclass[10pt,journal,compsoc]{IEEEtran}

\ifCLASSOPTIONcompsoc
  \usepackage[nocompress]{cite}
\else
  \usepackage{cite}
\fi

\usepackage[pagebackref=true,breaklinks=true,bookmarks=false]{hyperref}
\usepackage{color}
\usepackage{amsmath}
\usepackage{array}

\usepackage{graphicx}
\usepackage{algorithm}
\usepackage{algorithmic}
\usepackage{booktabs}
\usepackage{multirow}
\usepackage{amsfonts}
\usepackage{amssymb}

\usepackage{nicefrac}

\newcommand{\R}{\mathbb{R}}
\newcommand{\E}{\mathbb{E}}

\makeatletter
\newcommand{\ssymbol}[1]{$^{\@fnsymbol{#1}}$}
\makeatother

\begin{document}
%
\title{Aligning Source Visual and Target Language Domains for Unpaired Video Captioning}
%
%
%
%

\author{Fenglin~Liu,
        Xian~Wu,
        Chenyu~You,
        Shen~Ge,
        Yuexian~Zou,
        and~Xu~Sun

\IEEEcompsocitemizethanks{\IEEEcompsocthanksitem Fenglin~Liu and Yuexian~Zou are with ADSPLAB, School of ECE, Peking University. Yuexian~Zou also with the Peng Cheng Laboratory, Shenzhen, China. E-mail: \{fenglinliu98, zouyx\}@pku.edu.cn
\IEEEcompsocthanksitem Xu~Sun is with MOE Key Laboratory of Computational Linguistics, School of EECS, Peking University, China.
E-mail: xusun@pku.edu.cn
\IEEEcompsocthanksitem Xian~Wu and Shen~Ge are with Tencent, China.\protect\\
E-mail: \{kevinxwu, shenge\}@tencent.com
\IEEEcompsocthanksitem Chenyu~You is with Department of Electrical Engineering, Yale University, USA. E-mail: chenyu.you@yale.edu
}
\thanks{Manuscript received 12 April 2021; revised 23 August 2021; accepted 16
November 2021. Date of publication 2 December 2021}
\thanks{(Corresponding authors: Xu Sun, Yuexian Zou and Xian Wu.)}
\thanks{Recommended for acceptance by K. Saenko.}
\thanks{Digital Object Identifier no. 10.1109/TPAMI.2021.3132229}
}

%
%

\markboth{IEEE TRANSACTIONS ON PATTERN ANALYSIS AND MACHINE INTELLIGENCE, VOL. 44, NO. 12, DECEMBER 2022}%
{Shell \MakeLowercase{\textit{et al.}}: Bare Advanced Demo of IEEEtran.cls for IEEE Computer Society Journals}
%



\IEEEtitleabstractindextext{%
\begin{abstract}
Training supervised video captioning model requires coupled video-caption pairs. However, for many targeted languages, sufficient paired data are not available. To this end, we introduce the unpaired video captioning task aiming to train models without coupled video-caption pairs in target language. To solve the task, a natural choice is to employ a two-step pipeline system: first utilizing video-to-pivot captioning model to generate captions in pivot language and then utilizing pivot-to-target translation model to translate the pivot captions to the target language. However, in such a pipeline system, 1) visual information cannot reach the translation model, generating visual irrelevant target captions; 2) the errors in the generated pivot captions will be propagated to the translation model, resulting in disfluent target captions. To address these problems, we propose the Unpaired Video Captioning with Visual Injection system (UVC-VI). UVC-VI first introduces the Visual Injection Module (VIM), which aligns source visual and target language domains to inject the source visual information into the target language domain. Meanwhile, VIM directly connects the encoder of the video-to-pivot model and the decoder of the pivot-to-target model, allowing end-to-end inference by completely skipping the generation of pivot captions. To enhance the cross-modality injection of the VIM, UVC-VI further introduces a pluggable video encoder, i.e., Multimodal Collaborative Encoder (MCE). The experiments show that UVC-VI outperforms pipeline systems and exceeds several supervised systems. Furthermore, equipping existing supervised systems with our MCE can achieve 4\% and 7\% relative margins on the CIDEr scores to current state-of-the-art models on the benchmark MSVD and MSR-VTT datasets, respectively.

\end{abstract}

}

\maketitle

\IEEEdisplaynontitleabstractindextext


\IEEEpeerreviewmaketitle

\ifCLASSOPTIONcompsoc
\IEEEraisesectionheading{\section{Introduction}\label{sec:introduction}}
\else
\section{Introduction}
\label{sec:introduction}
\fi

\IEEEPARstart{V}{ideo} captioning targets to understand the visual content of given videos and generate corresponding descriptive sentences. Video captioning has a wide range of applications, such as video retrieval \cite{Yu2017Retrieval}, human-robot interaction \cite{das2017visual} and visually impaired people aiding \cite{Voykinska2016helpsee}. Due to its broad usage scenarios, video captioning has received extensive research interests. Among existing approaches, the encoder-decoder based systems \cite{Venugopalan2015vc3,Zheng2020SAAT,liu2021o2na,Ryu2021Semantic} have achieved great success in advancing the state-of-the-art. 

Currently, training a supervised video captioning model requires large volume of video and caption pairs. However, for many targeted languages (i.e., Non-English), sufficient video-caption pairs are not available. For example, to generate video captions in Chinese, it's a necessity to collect video-Chinese caption pairs which is both costly and time-consuming.
To this end, we introduce the problem of unpaired video captioning in which the video-caption pairs in target language are not available, which has practical value for non-English languages.

As shown in the left sub-figure of Figure~\ref{fig:model}, to build video captioning model without paired data, which has not been well studied yet, an intuitive approach is to build a pivot language based, two-step pipeline system. In implementation, given an input video, the pipeline system firstly generates the captions in pivot language by a video-to-pivot video captioning model; then translates the pivot captions to the target language by a pivot-to-target translation model. 
In this way, the pipeline system can leverage existing resources, e.g., video-pivot paired dataset (denoted as $D_1$), pivot-target paired dataset (denoted as $D_2$), as well as pre-trained video captioning and translation models.
Take video Chinese captioning for example, the pipeline system uses English as a pivot language, and trains a video-to-pivot captioning model on video-English pairs \cite{Xu2016MSR-VTT} and a pivot-to-target translation model on English-Chinese pairs (e.g. WMT En-Zh (\url{http://statmt.org/wmt17/})). Sending the video through the captioning model and subsequently the translation model, a video caption in Chinese can be generated.

The pipeline systems have achieved great success in many tasks, such as machine listening comprehension \cite{Tseng2016MRC}, spoken question answering \cite{Shiang2014Spoken} and unpaired machine translation \cite{wu2007pivotNMT}.
However, when applying to video captioning, the training of the captioning model and the translation model are rather separated compared to the conventional end-to-end systems.
As shown in Figure~\ref{fig:model}, on one hand, visual information in the video-to-pivot captioning model is unable to reach the pivot-to-target translation model, resulting in the lack of visual details in generated target captions, bringing in \textit{visual irrelevancy errors}; On the other hand, the errors in the pivot captions generated by the video-to-pivot captioning model cannot be corrected by the pivot-to-target translation model~\cite{samy2015scheduled,Vinyals2017Lessons}, resulting in \textit{disfluent target captions}.
Moreover, the optimization over the pivot-to-target translation model can not be back-propagated to the video-to-pivot captioning model.

To address above problems of the pipeline system, we propose the Unpaired Video Captioning with Visual Injection system (UVC-VI) which includes two major components: Visual Injection Module (VIM) and Multimodal Collaborative Encoder (MCE). As shown in the right sub-figure of Figure \ref{fig:model}, UVC-VI uses VIM and MCE to directly bridge the encoder of the captioning model and the decoder of the translation model.
As a result, 1) the source visual information can be injected into the target language domain, which addresses the visual irrelevancy problems;
2) UVC-VI no longer generates captions in pivot language, which addresses the disfluency problems by allowing end-to-end inference.
Therefore, our UVC-VI can solve the problems in the pipeline systems, and generate more detailed, accurate and fluent captions in target language.

For clarity, we first introduce the VIM, followed by the MCE.
The input of VIM is the visual embedding, which is the output of video encoder and has rich source visual information. Next, VIM uses the proposed pseudo supervised training and adversarial training to project the visual embedding to textual embedding, which injects the source visual information into the target language domain.
The textual embedding, i.e., projected visual embedding, is directly fed into the decoder part of the translation model to generate captions in target language.
We further propose the MCE to enhance the cross-modality projection of the VIM.
In detail, MCE first accepts original visual embedding from the video encoder and incorporates textual concepts\footnote{\textbf{Motivation}: Textual concepts contain a set of words describing \textit{object} (e.g., {cat}), \textit{attribute} (e.g., {small}) and \textit{relationship} (e.g., {standing}) of videos \cite{Pan2017Attributes}. Therefore, textual concepts provide a more semantic representation of visual information and thus help shorten the gap between visual and the language domains.} \cite{Pan2017Attributes} into them. The transformed textual-enriched visual embedding helps to shorten the modality gap between source visual and target language domains for VIM, and in turn enhance the cross-modality projection of the VIM.

In the absence of video-caption pairs in the target language domain  (denoted as $D_3$), we train our approach using existing resources, i.e., video-pivot paired dataset $D_1$ and pivot-target paired dataset $D_2$.\footnote{The two paired datasets, i.e., $D_1$ and $D_2$, can have no overlap.} To learn the parameters of VIM and MCE, we introduce two training approaches: pseudo supervised training and adversarial training.
Take video Chinese captioning for example, given video-English pairs $D_1$ and English-Chinese pairs $D_2$, 1) for the pseudo supervised training, we feed video and English caption pairs from $D_1$ into the encoder of the video-to-English captioning model and the encoder of English-to-Chinese translation model, to generate visual embedding ${{V}}$ and textual embedding ${{T}_1}$. The ${{V}}$-${{T}_1}$ pairs are used as pseudo paired data to train the VIM; 2) for the adversarial training, we further use $D_2$ by feeding the English sentence from $D_2$ into the encoder of English-to-Chinese translation model to generate textual embedding ${{T}_2}$. Since there are no pairs of $V$ and $T_2$ , we adopt the adversarial training \cite{goodfellow2014GAN,Zhu2017unpairedI2I} to exploit ${{V}}$ and ${{T}_2}$ to further train our VIM.
In this manner, once the VIM are trained, we can project (i.e., inject) the source visual information to the target language domain directly.
Besides, as shown in the right sub-figure of Figure~\ref{fig:model}, the encoder of the video-to-English captioning model and the decoder of the English-to-Chinese translation model can be concatenated by VIM to form an end-to-end video Chinese captioning model. The decoder of the captioning model and the encoder of the translation model are simply dropped away, so the pivot video captions in English are no longer generated.
In this way, our UVC-VI is capable of generating desirable and fluent captions without the training on the pairs of video and target caption $D_3$.
The experiments and analyses on two benchmark datasets, i.e., MSVD \cite{Guadarrama2013MSVD} and MSR-VTT \cite{Xu2016MSR-VTT}, prove our arguments and verify the effectiveness of our proposed approach.
Moreover, we extend our UVC-VI to image captioning task, obtaining positive experimental results.

Overall, our main contributions are as follows:

\begin{itemize}
    \item In this work, we introduce the problem of unpaired video captioning where the video-caption pairs in target language are not available, while all existing studies need sufficient video-caption pairs in target language.
    
    \smallskip\item We make the first attempt to conduct unpaired video captioning. In particular, we propose the Unpaired Video Captioning with Visual Injection system (UVC-VI), which consists of the Visual Injection Module (VIM) and the Multimodal Collaborative Encoder (MCE), to address the visual irrelevancy and disfluency problems in the pipeline systems.
    
    \smallskip\item The experiments show that the UVC-VI outperforms both pipeline systems and several supervised systems over all metrics. In addition to automatic metrics, we also conduct human evaluations on three different target languages, i.e., Chinese, French and German, to verify our arguments and prove the advantage of our proposed UVC-VI from the perspective of user experience.
    
    \smallskip\item It is worth noting that the MCE can be easily integrated into conventional supervised video captioning systems as an individual video encoder module. Equipping MCE helps these systems to achieve new state-of-the-art results on MSVD and MSR-VTT benchmark datasets.
    
\end{itemize}

\section{Related Work}

\subsection{Conventional Supervised Video Captioning}
In recent years, there is a surge of research interests in video captioning.
There are a large number of encoder-decoder based neural models proposed for video captioning \cite{Venugopalan2015vc3,Xu2017MA-LSTM,Pei2019MARN,Chen2019Temporal,Aafaq2019GRU-EVE,Ryu2021Semantic,Zhang2020ORG,liu2021o2na}.
State-of-the-art approaches \cite{Pei2019MARN,Chen2019MGSA,liu2020prophet,Chen2019Temporal,Ryu2021Semantic} introduce a video encoder to encode the video and a language decoder, e.g., LSTM \cite{hochreiter1997long} and Transformer \cite{Vaswani2017transformer,Zhou2018Transformer,liu2021o2na}, to generate coherent captions with the attention mechanism \cite{bahdanau2014neural,Pan2016vc1}.
However, all of existing video captioning models are trained on large-scale video-caption pairs, while collecting paired video-caption data for training is expensive and time-consuming.
In contrast to the existing models, we utilize the pipeline system to relax the reliance on the paired dataset for video captioning.

\subsection{Pipeline Systems}
Currently, pipeline systems have been proved to be effective in various applications: 1) Machine Listening Comprehension \cite{Tseng2016MRC} and Spoken Question Answering \cite{Shiang2014Spoken} which include an automatic speech recognition (ASR) model \cite{Yu2017ASR} followed by either a machine comprehension model \cite{Lai2017RACE} or an information retrieval model \cite{Shiang2014Spoken};
2) Medical Image Classification \cite{Wong2018medical} which includes a image segmentation model \cite{Mehta2017m_net} followed by a classification model;
3) Unpaired Machine Translation \cite{wu2007pivotNMT,Utiyama2007PivotComparison,Zahabi2013Improving} which stacks a source-to-pivot translation model followed by a pivot-to-target translation model.
Specifically, in image captioning, \cite{wu2016what} first adopts a multi-label classification framework to predict the visual attributes, and then employs an RNN model to generate captions. 
Although these pipeline systems achieve success in multiple tasks, stacking multiple models can cause error accumulation problems. For example, in machine listening comprehension, \cite{Lee2019Machine} found that ASR errors severely impair the machine comprehension system. 
Similarly,  we find that the pipeline systems for unpaired video captioning will bring the visual irrelevancy and disfluency errors.

\subsection{Unpaired Image Captioning}
\label{sec:UIC}
Although the unpaired video captioning has not been well studied yet, the unpaired image captioning has been explored recently \cite{LanLD17MM,Song2019Unpaired,gu2018unpaired,gu2019unpaired,Feng2019Unsupervised_ic,liu2019unpaired,Laina2019Towards,Yang2020USGAE}.
Specifically, unpaired image captioning, which is also known as unpaired image-to-sentence translation, is essentially similar to the problem of unpaired sentence-to-sentence translation \cite{wu2007pivotNMT,Utiyama2007PivotComparison} and unpaired image-to-image translation \cite{Zhu2017unpairedI2I}.
Typically, the source sentence/image and the target sentence/image are mapped into a common latent space, in which sentences/images with the same semantic/visual meaning are well aligned so that unpaired translation can be performed.
Nevertheless, due to the great disparities between the vision and the language domains, unpaired image captioning is considerably more challenging.

To train unpaired image captioning models, existing approaches can be classified into two categories: unsupervised approaches and pipeline approaches.
For the unsupervised approaches, \cite{gu2019unpaired,Yang2020USGAE} and \cite{Feng2019Unsupervised_ic,Laina2019Towards} take the scene graph and textual concepts to align the vision and the language domains, respectively, in an adversarial way.
Although these works are totally unsupervised, the existing resources like large-scale paired datasets and pre-trained models are not fully utilized. 
For the pipeline approaches, \cite{gu2018unpaired} first adopts a pivot-based model to generate captions in a pivot language (Chinese) and then translate them into the target language (English). 
In particular, they improve pipeline approaches by adapting 1) the encoder of the translation model to the decoder of the captioning model; 2) the decoder of the translation model to the decoder of a pre-trained auto-encoder. Since only text embedding parameters are optimized and more processing steps are appended, the drawbacks of pipeline approaches, i.e., visual irrelevancy and disfluency errors, are not fully addressed.
Besides, \cite{LanLD17MM} and \cite{Song2019Unpaired} also introduce the translation model in unpaired image captioning, but they did not follow the pipeline approach. Instead they translate the captions in pivot language to the target language, and use the translated captions to train the image to target language captioning model.
Since the translation model is fixed in advance, the translation error could constantly affect the captioning performance.

Although unpaired image captioning has been explored, the unpaired video captioning is relatively more challenging, because videos involve scenes that are volatile and likely to change.
Besides, there are three source modalities (i.e., image, motion and audio) and the temporal dynamics information should be captured to understand the video efficiently, while image captioning only use image as input.
In our work, we focus on making full use of existing resources, e.g., video-pivot and pivot-target paired datasets and pre-trained models, to perform the unpaired video captioning.

\begin{figure*}[t]
\centering
\includegraphics[width=0.495\linewidth]{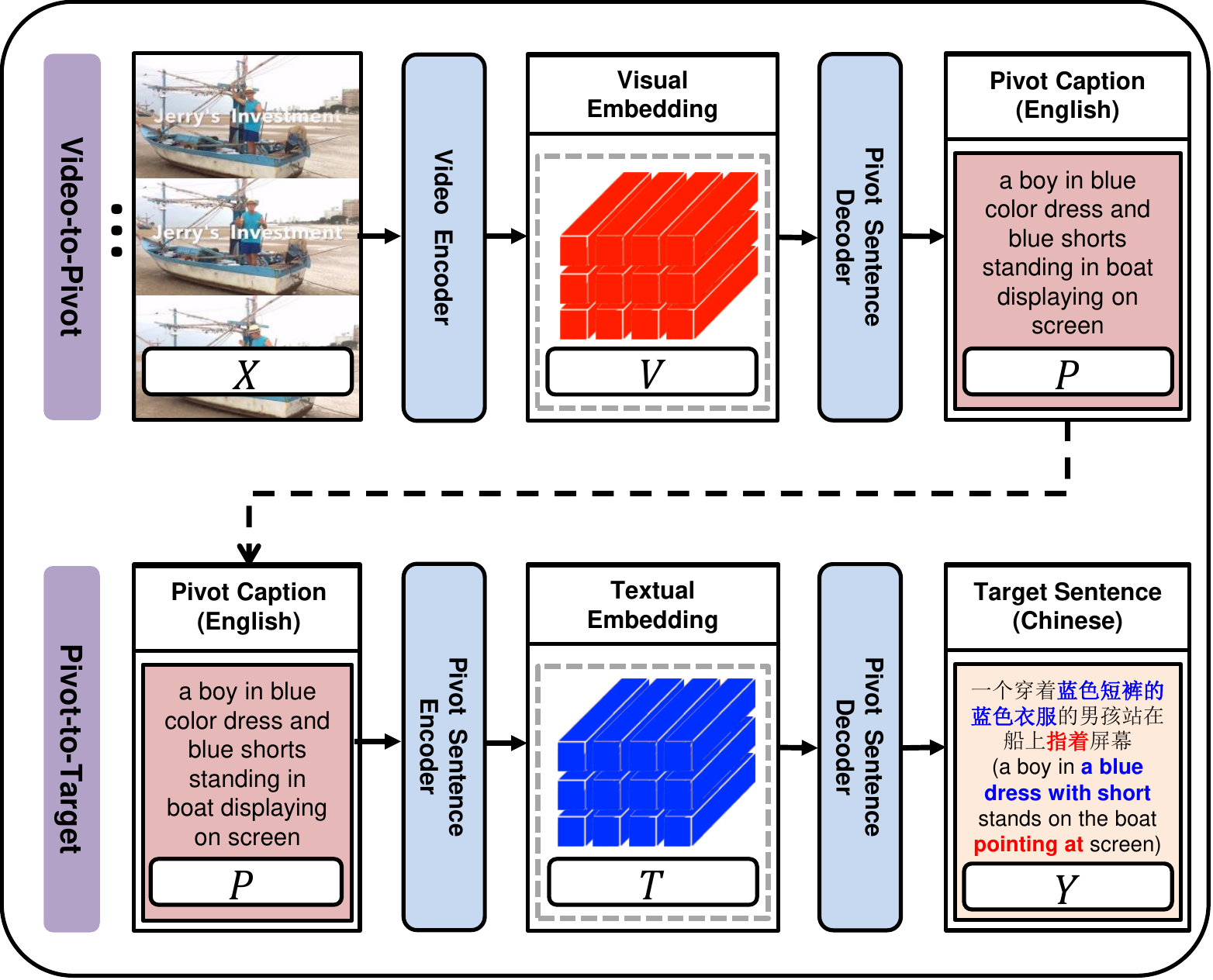}
\hfil
\includegraphics[width=0.495 \linewidth]{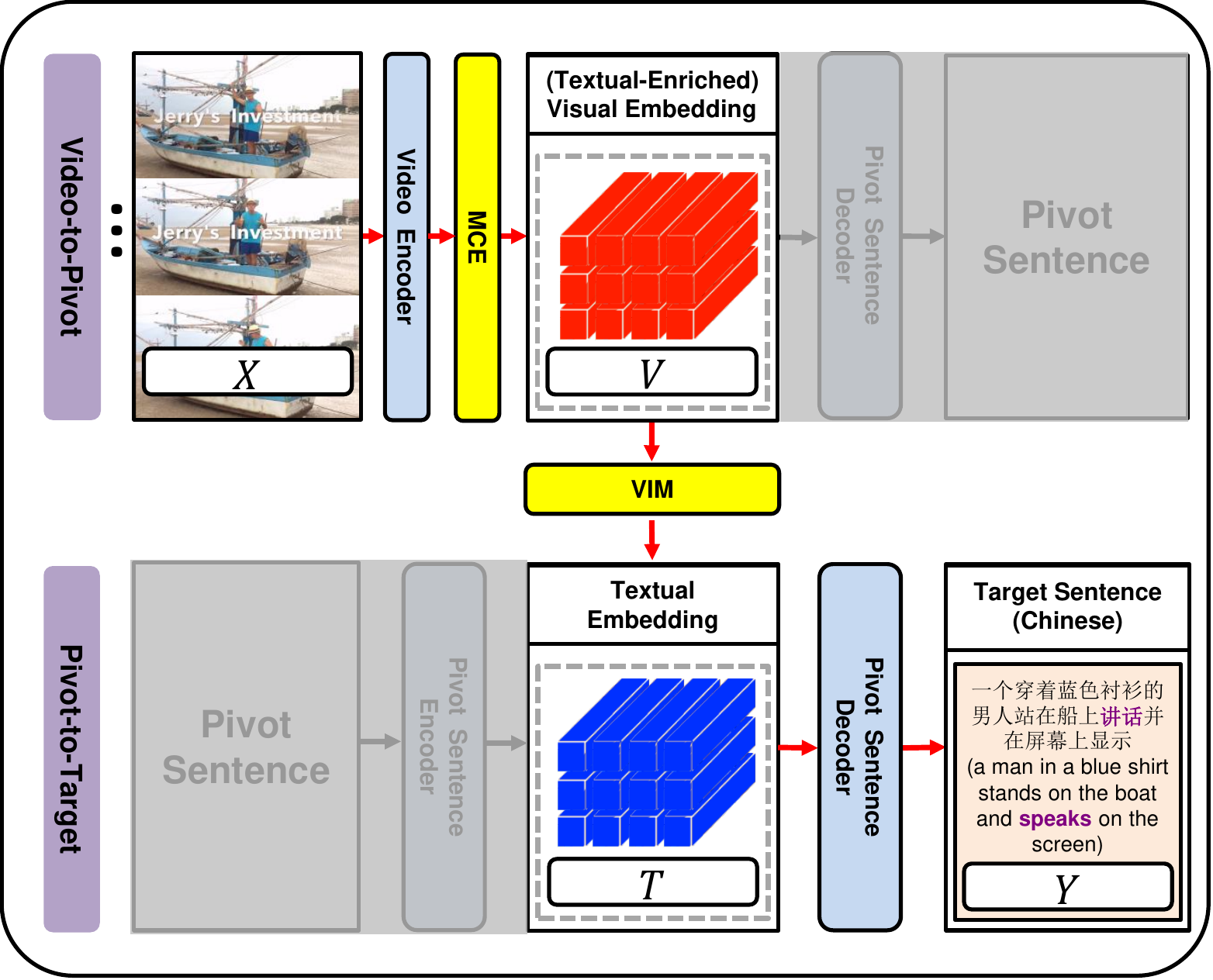}

\caption{Illustration of the pipeline system (left) and our UVC-VI system (right) under the video-to-Chinese captioning scenario. The target caption generated by pipeline system suffers from visual irrelevancy (see words highlighted red) and disfluency (see words highlighted blue) errors. The words highlighted purple denote the detailed visual element that is included in the caption. For better understanding, we add the English translation below the Chinese captions in brackets.}
\label{fig:model}
\end{figure*}

\section{Approach}
We first formulate the conventional supervised video captioning problems; Then, we describe the pipeline system and the proposed UVC-VI for unpaired video captioning.

\subsection{Problem Formulation of Video Captioning}
\label{sec:formulation}

Given a video ${X}$, the goal is to generate a descriptive target sentence ${Y}=\{y_1,y_2,\dots,y_T\}$.
The state-of-the-art systems \cite{Venugopalan2015vc3,Pei2019MARN,Chen2019MGSA,Chen2019Temporal} normally include a video encoder and a target sentence decoder, which can be formulated as:
\begin{align}
\text{Video Encoder}  : X \to V ; \ \
\text{Target Decoder} : V \to Y .
\label{eqn:caption}
\end{align}
The video encoder aims to extract the visual embedding ${{V}}$ of input video. We use the set ${{V}} = [{{V}}_\text{T}; {{V}}_\text{I}; {{V}}_\text{M}; {{V}}_\text{A}]$ to denote the visual embedding of video. ${{V}}_\text{I} \in \R^{N \times d_i}$ denotes the image features extracted by CNNs \cite{Szegedy2017Inception-v4}; ${{V}}_\text{M} \in \R^{N \times d_m}$ and  ${{V}}_\text{A} \in \R^{N \times d_a}$ denote the motion and audio features in video which can be extracted by 3D CNN \cite{Tran2015C3D} and Bag-of-Audio-Words \cite{Pancoast2014boaw}, respectively.
In this paper, we further follow \cite{Pan2017Attributes} to extract a set of textual concepts ${{V}}_\text{T} \in \R^{N \times d_c}$ from the input video, which describes the \textit{object} (e.g., table), \textit{attribute} (e.g., wooden) and \textit{relationship} (e.g., sitting) of the video. The embedding vectors of these textual concepts are then taken as the ${{V}}_\text{T}$. More details of how to generate ${{V}}_\text{T}$ can be found in \cite{Pan2017Attributes}.
In implementation, all extracted features are projected to the same dimension of $d_h$, constituting the visual embedding ${{V}} = [{{V}}_\text{T}; {{V}}_\text{I}; {{V}}_\text{M}; {{V}}_\text{A}] \in \R^{4N \times d_h}$.

The language decoder, e.g., LSTM \cite{hochreiter1997long} and Transformer \cite{Vaswani2017transformer}, is used to predict the descriptive target sentence ${Y}$ from ${{V}}$.
Given the ground truth sentence for the source video, we can simply train an encoder-decoder system by minimizing supervised training loss, e.g., cross-entropy loss.

Here, we introduce three datasets: 1) the video-pivot paired dataset $D_1$ ; 2) the pivot-target paired dataset $D_2$ ; 3) the video-target paired dataset $D_3$.
For example, in the context of generating Chinese captions for input videos, $D_1$ refers to pairs of videos and their corresponding captions in English; $D_2$ refers to pairs of English sentences and their translations in Chinese; $D_3$ refers to pairs of videos and their corresponding captions in Chinese.\footnote{The English captions in $D_1$ and the English sentences in $D_2$ are separate sets and can have no overlap.}
To train the video Chinese captioning model in Eq.~(\ref{eqn:caption}), all existing methods need the dataset in the form of $D_3$. Although many datasets of video-English caption pairs ($D_1$) have been released \cite{Xu2016MSR-VTT,Zhou2018youcook2,Krishna2017ActyNet,Guadarrama2013MSVD}, in terms of non-English languages, only a very few datasets ($D_3$) \cite{Wang2019VATEX} are available.
As a result, insufficient paired data poses a great challenge for building video captioning models for non-English languages.

\subsection{Pipeline System for Unpaired Video Captioning}
\label{sec:pipeline}
A simple yet effective solution is to build a pipeline system, assembling a video-to-pivot captioning model and a pivot-to-target translation model together. The trainings are conducted independently on different datasets, i.e., a video-pivot paired dataset $D_1$ for training the video-to-pivot video captioning model, as well as a pivot-target paired dataset $D_2$ for training the pivot-to-target machine translation model. As a result, there is no need for a video-target paired dataset $D_3$.
As shown in the left sub-figure of Figure~\ref{fig:model}, the pipeline system can be formulated as:
\begin{equation}
\begin{aligned}
&\text{Video Encoder}  : X \to V ;
\quad
\text{Pivot Decoder} : V \to P ; 
\\
&\text{Pivot Encoder} : P \to T ;
\quad
\text{Target Decoder} : T \to Y ,
\label{eqn:stack}
\end{aligned}
\end{equation}
where ${P}$ denotes the pivot sentence generated by the video-to-pivot captioning model and ${{T}}$ denotes the textual embedding in the target language.
As we can see, the visual embedding $V$, which contains the rich visual information, are separated from the translating process by the intermediate pivot caption, leading to possible visual irrelevancy errors. 
Moreover, compared with Eq.~(\ref{eqn:caption}), two more steps are added to Eq.~(\ref{eqn:stack}), which increases the chances of error generation and amplification, especially the generation process \cite{samy2015scheduled,Vinyals2017Lessons}, leading to possible disfluency errors.

\begin{table}[t]
\centering
\footnotesize
\caption{Training datasets needed for different approaches. $D_1$, $D_2$ and $D_3$ represent the video-pivot paired dataset, pivot-target paired dataset and video-target paired dataset, respectively. There is no overlap between $D_1$, $D_2$ and $D_3$.
As we can see, our approach aims to leverage existing resources, e.g., video-pivot and pivot-target paired datasets, to perform the unpaired video captioning, i.e., generate fluent and desirable captions in target language without the training on video-target paired dataset $D_3$. 
As a result, our approach has practical value for non-English languages, especially for the low-resource target language applications, e.g., French and German, in which the pairs of video and target caption are not available.}
\label{tab:datasetfortraining}
\begin{tabular}{@{}l c c c@{}}
\toprule
{Methods}
& {$D_1$}
& {$D_2$}   
& {$D_3$} \\ \midrule [\heavyrulewidth]
Conventional Supervised System. i.e., Eq.~(\ref{eqn:caption})  & $\times$ & $\times$ & $\surd$ \\
Pipeline System, i.e., Eq.~(\ref{eqn:stack}) & $\surd$ & $\surd$ & $\times$ \\
Proposed UVC-VI System, i.e., Eq.~(\ref{eq:uvc-vi}) & $\surd$ & $\surd$ & $\times$ \\
\bottomrule
\end{tabular}
\end{table}

\subsection{Unpaired Video Captioning with Visual Injection}
\label{sec:uvcvi}
To address the issues in the pipeline systems, we propose the UVC-VI system, which introduces the Multimodal Collaborative Encoder (MCE) and Visual Injection Module (VIM) to connect the encoder of the video-to-pivot captioning model and the decoder of the pivot-to-target translation model. As shown in the Figure~\ref{fig:model}, our proposed UVC-VI can be formulated as:
\begin{equation}
\begin{aligned}
\label{eq:uvc-vi}
& \text{Video Encoder + MCE} : X \to {{V}} ;
\\  \text{VIM} &: {{V}} \to {{T}} ;
\quad
\text{Target Decoder} : {{T}} \to {Y} .
\end{aligned}
\end{equation}
Compared to Eq.~(\ref{eqn:stack}), in Eq.~(\ref{eq:uvc-vi}), we make three updates: 1) injecting the source visual information into the target language; 2) replacing two steps, i.e., Pivot Decoder and Pivot Encoder, in Eq.~(\ref{eqn:stack}) with the proposed MCE and VIM; 3) skip the generation of pivot captions ${P}$.
Specifically, to efficiently inject the source visual information into the target language, i.e., conduct the cross-modality projection, VIM should devote on aligning source visual and target language domains \textit{without the training on the pairs of video and target caption}.
Then, the MCE is proposed to enhance the cross-modality projection of the VIM.
As a result, 1) the source visual information can be injected into the target language domain, which addresses the visual irrelevancy errors;
2) UVC-VI no longer generates captions in pivot language, allowing end-to-end inference, which addresses the disfluency errors.

For clarity, we first introduce the VIM in Section~\ref{sec:VIM}, then introduce the MCE in Section~\ref{sec:MCE}. 
As shown in Eq.~(\ref{eq:uvc-vi}), VIM first receives visual embedding $V$, and then projects the visual embedding $V$ to textual embedding $T$, i.e., inject the source visual information into the target language domain.
Therefore, the input of VIM is ${V}$, the output of VIM is ${T}$.
Due to the lack of video-target paired dataset $D_3$, i.e., $X$-$Y$ pairs, we can not train our approach by minimizing supervised training loss as like in existing works. 
Instead, we propose the pseudo supervised training and adversarial training to train the cross-modality projection of VIM. 
In implementations, as shown in Table~\ref{tab:datasetfortraining}, we make full use of the video-pivot paired dataset $D_1$ and the pivot-target paired dataset $D_2$.
Firstly, we feed the video and pivot caption pairs from $D_1$ into the encoder of the video-to-pivot captioning model and the encoder of pivot-to-target translation model, respectively, to acquire visual embedding ${{V}}$ and textual embedding ${{T}_1}$.  The ${{V}}$-${{T}_1}$ pairs can be used as pseudo paired data in pseudo supervised training to train the VIM.
Secondly, we further feed the pivot sentence from $D_2$ into the encoder of pivot-to-target translation model to acquire the textual embedding ${{T}_2}$.
Since the two datasets, i.e., $D_1$ and $D_2$, can have no overlap, there are no pairs of ${{V}}$ and ${{T}_2}$. To exploit the $V$ and ${{T}_2}$ to improve the performance, we introduce the adversarial training \cite{goodfellow2014GAN} to further train the proposed VIM.

\subsubsection{Visual Injection Module (VIM)}
\label{sec:VIM}

As expected, the VIM aims to inject the visual information into the target language domain by aligning the source visual and target language domains.
In other words, the VIM devotes on projecting the visual embedding ${V}$ of video domain to the textual embedding ${{T}}$ of target language domain without the need for video-target pairs.
To achieve this, we empirically find that a multilayer perceptron (MLP) \cite{rumelhart1985learning} can efficiently conduct the projection, therefore, we directly apply the MLP to implement the VIM:
\begin{equation}
{{T}} \approx \text{VIM}(V) = \max\left(0,V{W}_\text{f}+{b}_\text{f}\right){W}_\text{ff}+{b}_\text{ff}
\label{eqn:vim} ,
\end{equation}
where ${W}_\text{f} \in \R^{d_h \times 2d_h}$ and ${W}_\text{{ff}} \in \R^{2d_h \times d_h}$ are linear transformation matrices; ${b}_\text{f}$ and ${b}_\text{ff}$ are the bias terms.

\smallskip\noindent\textbf{Parameter Optimization} \ As mentioned above, in order to estimate the parameters of VIM, we propose two training methods, pseudo supervised training and adversarial training. The pseudo supervised training method relies on the video-pivot pairs from $D_1$ for training and the adversarial training method further relies on the pivot-target pairs from $D_2$ for training. It is worth noting that neither method requires the pairs of video and caption in target language.

$\bullet$ \noindent\textbf{Pseudo Supervised Training} $\quad$
In order to effectively make use of $D_1$, we propose the pseudo supervised training.
In implementation, we can generate a large number of the pairs between ${{V}}$ and ${{T_1}}$ by running the encoder part of video-to-pivot captioning model and the encoder part of pivot-to-target translation model on the video-pivot pairs from $D_1$.
In detail, we first extract the visual embedding ${{V}}$ by inputting the video from $D_1$ into the encoder of the captioning model;  Then we extract the textual embedding ${{T_1}}$ by inputting the pivot sentence from $D_1$ into the encoder of the translation model. 
In this way, we can acquire the coupled ${{V}}$-${{T_1}}$ pairs.
Since UVC-VI aims to project the visual embedding ${{V}}$ to the textual embedding ${{T_1}}$, we use the L1 norm between $\text{VIM}(V)$ and $T_1$ as the training loss:
\begin{align}
{L}_{\text{pseudo}}({{V}}, {{T_1}}) = \E_{({{V}}, {{T_1}}) \sim D_1}\left[\| \text{VIM}\left(V\right) - {{T_1}}\|_{1}\right] .
\end{align}

$\bullet$ \noindent\textbf{Adversarial Training} $\quad$
In order to effectively make use of the pivot-target paired dataset $D_2$, we further extract the textual embedding $T_2$ by inputting the pivot sentence from $D_2$ into the encoder of the translation model. 
Based on the visual embedding $V$ from $D_1$ and the textual embedding $T_2$ from $D_2$, since the two datasets, i.e., $D_1$ and $D_2$, can have no overlap, there are no pairs of $V$ and $T_2$, so we introduce the adversarial training \cite{Zhu2017unpairedI2I,goodfellow2014GAN} to train our approach.
In implementation, we introduce a discriminator $\text{D}$ to distinguish between the $\text{VIM}(V)$ and the $T_2$. Then we adopt the adversarial loss in Eq.~(\ref{eq:gan1}):
\begin{equation}
\begin{aligned} 
\label{eq:gan1}
{L}_\text{adv}({{V}}, {{T_2}})  = \
& \E_{{{V}} \sim D_1}\left[\log\left(1-\text{D}(\text{VIM}({{V}}))\right)\right] \\
& + \E_{{{T_2}} \sim D_2}\left[\log \text{D}({{T_2}})\right] .
\end{aligned}
\end{equation}
With the adversarial loss, the VIM devotes on projecting the visual embedding $V$ of video domain to target language domain $\text{VIM}(V)$, while $\text{D}$ devotes on distinguishing between projected visual embedding $\text{VIM}(V)$ and textual embedding $T_2$.
In other words, through the adversarial training, we could align the latent space of video domain and target language domain without the requirement for video-target pairs.

Inspired by \cite{Zhu2017unpairedI2I}, we further introduce a cycle strategy to enhance such alignment.
For the cycle strategy, a Textual Injection Module (TIM), which shares the same structure as VIM and is defined as:
\begin{equation}
V \approx \text{TIM}(T) = \max\left(0,T{W}'_\text{f}+{b}'_\text{f}\right){W}'_\text{ff}+{b}'_\text{ff} ,
\end{equation}
and a new discriminator ${\text{D}'}$ are introduced, where the former devotes on injecting textual embedding ${{T_2}}$ to the visual domain and the latter devotes on discriminating between $\text{TIM}(T_2)$ and the $V$.
Similarly as Eq.~(\ref{eq:gan1}), there is an adversarial loss ${{L}'}_\text{adv}(T_2,V)$ to train the TIM and ${\text{D}'}$:
\begin{equation}
\begin{aligned}
\label{eq:gan-reverse}
{{L}'}_\text{adv}({{T_2}},{{V}})  = & \E_{{{T_2}} \sim D_2 }\left[\log\left(1-{\text{D}'}(\text{TIM}({{T_2}}))\right)\right] \\
& + \E_{{{V}} \sim D_1}\left[\log {\text{D}'}(V))\right] .
\end{aligned}
\end{equation}
The cycle consistency loss is further introduced to regularize the adversarial training \cite{Zhu2017unpairedI2I}:
\begin{equation}
\begin{aligned}
{L}_{\text{cyc}}({{V}}, {{T_2}})
= & \E_{{{V}} \sim D_1 }\left[\|\text{TIM}(\text{VIM}(V))) - V)\|_{1}\right] + \\
& \E_{{{T_2}} \sim D_2 }\left[\|\text{VIM}(\text{TIM}({{T_2}})) - {{T_2}}\|_{1}\right] .
\end{aligned}
\end{equation}
With above equations, the proposed UVC-VI is able to maintain the backward cycle consistency, i.e., $V \to \text{VIM}(V) \to \text{TIM}(\text{VIM}(V))\approx V$.

Overall, combining the pseudo supervised training loss and adversarial training loss, the full training objective is defined as:
\begin{align}
\label{eq:full_loss}
{L}_{\text{full}}(V, & \ T_1, T_2)
=  {L}_{\text{pseudo}}(V, T_1) \ + \nonumber  \\
& \alpha\left({L}_{\text{adv}}({{V}}, T_2) + {L}'_{\text{adv}}(T_2, V)  + \beta {L}_{\text{cyc}}(V, T_2)\right) ,
\end{align}
where $\alpha = 0.1$ and $\beta = 10$ are the hyper-parameters that control the regularization.

Through the introduced Eq.~(\ref{eq:full_loss}), we are able to estimate the parameters of the VIM.
As a result, the VIM can align the source visual and target language domains, resulting in efficiently projecting the visual embedding of source video domain to the textual embedding of target language domain without the requirement for video-target pairs.

\subsubsection{Multimodal Collaborative Encoder (MCE)}
\label{sec:MCE}
In this section, we further introduce the Multimodal Collaborative Encoder (MCE) to enhance the cross-modality projection of the Visual Injection Module (VIM).
Since the visual embedding will be projected to the language domain and will be used for textual caption generation, based on the attention mechanism \cite{Vaswani2017transformer}, we propose the MCE to transform the original visual embedding into the textual-enriched visual embedding.

In implementation, as stated in Section~\ref{sec:formulation}, the ${{V}}_\text{T}$ denotes the set of textual concepts extracted from the video. According to the attention theorem\footnote{The attention mechanism computes the association weights between different features. Thus, the attention mechanism allows probabilistic many-to-many relations instead of monotonic relations, as in \cite{Vaswani2017transformer,xu2015show,fenglin2019mia}.}, if we take textual features ${{V}}_\text{T}$ as the query, and take the image/motion/audio features as the key and value, the image/motion/audio features will be associated with the textual features. 
Thus, through combining the textual features with the associated image/motion/audio features, we can acquire the textual-enriched image/motion/audio embedding, i.e., textual-enriched visual embedding.
At last, we further introduce a gate mechanism \cite{Cornia2020M2} to calibrate the contributions of each type of features. The MCE is defined as:
\begin{equation}
{V}' = \text{MCE}(V) 
= \sum_{i \in \{\text{I}, \text{M}, \text{A}\}} {\gamma}_{i} \odot \text{Attention}\left({{V}}_\text{T}, {{V}}_i\right) ,
\end{equation}
where $V'$ denotes the text-enriched visual embedding and the ${\gamma}_{i}$ and $\text{Attention}\left(\cdot,\cdot\right)$ are defined as follows:
\begin{equation}
\begin{aligned}
    {\gamma}_{i} = &  \sigma\left(\left[{{V}}_\text{T}; \text{Attention}\left({{V}}_\text{T}, {{V}}_i\right)\right]W_{i}\right) , \\
\text{Attention}&(x, y) = \text{softmax}\left(x W_\text{q}\left(y W_\text{k}\right)^{\top}\right) y W_\text{v} .
\end{aligned}
\end{equation}
In above equations, $\odot$ stands for the element-wise multiplication; $\sigma$ represents the sigmoid activation; and $[\cdot; \cdot]$ denotes the concatenation operation.
$W_{i} \in \R^{2d_h \times d_h}$, $W_\text{q} \in \R^{d_h \times d_h}$, $W_\text{k} \in \R^{d_h \times d_h}$ and $W_\text{v} \in \R^{d_h \times d_h}$ are learnable parameters.
The proposed MCE is then followed by dropout \cite{srivastava2014dropout}, shortcut connection \cite{he2016deep} and layer normalization \cite{ba2016layernormalization}.

The obtained textual-enriched visual embedding ${V}'$ can shorten the modality gap between visual and textual modalities for VIM.
In our UVC-VI, during the training of VIM, i.e., Eq.~(\ref{eq:full_loss}), we directly replace the $V$ with ${V}'$.
Meanwhile, the MCE can be integrated into other existing supervised video captioning models to improve their performance (see Section~\ref{sec:Analysis_MCE} and Table~\ref{tab:MCE}).

\begin{table*}[t]
    \centering
    \footnotesize  
    \caption{Performance of automatic evaluations on the MSVD and MSR-VTT video captioning datasets under the unpaired setting. The I, M and A denote image, motion and audio features, respectively. Higher value denotes better performance in all columns.
    The \underline{Underlined} numbers denote the best results of unpaired methods, where coupled video and target captions are not available; The \textbf{\underline{\underline{Bold}}} numbers denote the best results across all approaches, both supervised and unpaired.
    \ssymbol{3} denotes our own implementation.
    As we can see, both our UVC-VI (I+M+A) and UVC-VI (I+M) outperform the pipeline systems, even surpassing several supervised video captioning models in terms of CIDEr.
    \label{tab:unpaired_auto}}
    \begin{center}
    \setlength{\tabcolsep}{2pt}
    \begin{tabular}{@{}c l l c c c c c c c c@{}}
        \toprule
        \multirow{2}{*}[-3pt]{Types}& \multirow{2}{*}[-3pt]{Methods} & \multirow{2}{*}[-3pt]{Features} &  \multicolumn{4}{c}{Dataset: MSVD \cite{Guadarrama2013MSVD}}   &  \multicolumn{4}{c}{Dataset: MSR-VTT \cite{Xu2016MSR-VTT}}  \\ \cmidrule(lr){4-7} \cmidrule(lr){8-11} & & & BLEU-4 & METEOR & ROUGE-L & CIDEr & BLEU-4 & METEOR & ROUGE-L & CIDEr\\ \midrule [\heavyrulewidth]
        \multicolumn{11}{c}{Setting: Conventional Supervised Video Captioning}  \\ \midrule 
        \multirow{12}{*}{\begin{tabular}[c]{@{}c@{}} Supervised \\ Systems \\ (Section~\ref{sec:formulation}) \end{tabular}} 
        & AF$_\text{(ICCV2017)}$ \cite{Hori2017Attention} & I+M+A & - & - &  - & - & 39.7 &  25.5&  - & 40.0 \\
        & MA-LSTM$_\text{(ACMMM2017)}$ \cite{Xu2017MA-LSTM} & I+M+A & - & - &- & -  & 36.5&  26.5&  59.8&  41.0 \\
        & Two-stream$_\text{(TPAMI2020)}$ \cite{Gao2020Hierarchical} & I+M &\bf \underline{\underline{54.3}} &33.5 & - & 72.8 & 39.7 & 27.0 & - & 42.1 \\
        & PickNet$_\text{(ECCV2018)}$ \cite{Chen2018PickNet} & I & 52.3 & 33.3 &  69.6 & 76.5 & 39.4 & 27.3&  59.7&  42.3 \\
        &RecNet$_\text{(TPAMI2020)}$ \cite{Zhang2020Reconstruct}  & I & 52.3 & 34.1 & 69.8 & 80.3 & 39.1 & 26.6 & 59.3 & 42.7 \\
        & TDConvED$_\text{(AAAI2019)}$ \cite{Chen2019Temporal} & I & 53.3 & 33.8 & - & 76.4 & 39.5 & 27.5 & - & 42.8 \\
        & POS-CG$_\text{(ICCV2019)}$ \cite{Wang2019Controllable} & I+M & - & - & - & - & 38.3 & 26.8 & 60.1 & 43.4\\  
        & STAT$_\text{(TMM2020)}$ \cite{Yan2020STAT} & I+M & 52.0 & 33.3 & - & 73.8 & 39.3 & 27.1 & - & 43.8 \\
        & GRU-EVE$_\text{(CVPR2019)}$ \cite{Aafaq2019GRU-EVE} & I+M &  47.9 & 35.0 & 71.5 & 78.1 & 38.3 &28.4 & 60.7 & 48.1 \\  
        & SAAT$_\text{(CVPR2020)}$ \cite{Zheng2020SAAT} & I+M  & 46.5 & 33.5 & 69.4 & 81.0 & 40.5 &  28.2 & \bf \underline{\underline{60.9}} &  49.1 \\  
        &SGN$_\text{(AAAI2021)}$ \cite{Ryu2021Semantic} &  I+M &52.8 & \bf \underline{\underline{35.5}} & \bf \underline{\underline{72.9}} & \bf \underline{\underline{94.3}}& 40.8 & 28.3 & 60.8 & 49.5 \\
        
        & MGSA$_\text{(AAAI2019)}$ \cite{Chen2019MGSA}  & I+M+A &  - & - &- &- & \bf \underline{\underline{45.4}} & \bf \underline{\underline{28.6}} & - & \bf \underline{\underline{50.1}} \\ 
        \midrule [\heavyrulewidth]
        \multicolumn{11}{c}{Setting: Unpaired Video Captioning}  \\ \midrule
        
        \multirow{4}{*}{\begin{tabular}[c]{@{}c@{}} Pipeline \\ Systems \\  (Section \ref{sec:pipeline}) \end{tabular}} &MA-LSTM \cite{Xu2017MA-LSTM} + Google Translator \cite{Wu2016google}\ssymbol{3} & I+M & 43.1 & 29.5 & 65.8 & 53.3 & 31.5 & 23.9 & 54.2 & 30.6 \\
        
        &SAAT \cite{Zheng2020SAAT} + Google Translator \cite{Wu2016google}\ssymbol{3} & I+M & 46.4 & 30.2 & 66.4 & 61.1 & 35.2 & 25.8 & 57.2 & 37.8 \\
        
        &SGN \cite{Ryu2021Semantic} + Google Translator \cite{Wu2016google}\ssymbol{3} & I+M & \underline{50.7}  & 32.6 & 69.2 & 72.9 & 38.0 & 26.7 & 57.1 & 39.6 \\

        &Base Model & I+M(+A) & 47.2 & 31.8 & 67.6 & 68.9 & 34.7 & 25.1 & 55.8 & 36.3  \\
        
        \midrule 
        
        \multirow{2}{*}{\begin{tabular}[c]{@{}c@{}} Proposed \\  (Section \ref{sec:uvcvi}) \end{tabular}}  & \multirow{2}{*}{UVC-VI}  & I+M &     49.6 &   \underline{34.7} &   \underline{70.3} &  \underline{83.4} & 37.2 & 26.8 & 57.7 & 43.7 \\
        
        & & I+M+A & -  & -  & -  & - &    \underline{38.9} & \underline{27.8} & \underline{59.5} &  \underline{44.5}   \\

        \bottomrule
    \end{tabular}
    \end{center}
\end{table*}

\section{Experiments}
In this section, we firstly describe the implementation details, which consist of model settings and training details.
Then we present the evaluation of our proposed approach on the benchmark datasets.

\subsection{Implementation Details}
\subsubsection{Model Settings}
For the visual embedding ${{V}} = [{{V}}_\text{T}; {{V}}_\text{I}; {{V}}_\text{M}; {{V}}_\text{A}] \in \R^{4N \times d_h}$, which includes the textual features ${{V}}_\text{T}$, image features ${{V}}_\text{I}$, motion features ${{V}}_\text{M}$ and audio features ${{V}}_\text{A}$, given a video, $N=8$ key frames are uniformly sampled to extract image features $V_\text{I}$. Considering both the past and the future contexts, we take each key frame as the center to generate corresponding motion features $V_\text{M}$ and audio features $V_\text{A}$.
For the image features $V_\text{I}$, which are good at illustrating the shapes, the colors and the relationships of the items in the image, we adopt the Inception-ResNet-V2 \cite{Szegedy2017Inception-v4} pre-trained on the ImageNet \cite{Deng2009ImageNet} to extract the 1536-D CNN-based image features $V_\text{I} \in \R^{8 \times 1536}$. 
The motion features $V_\text{M} \in \R^{8 \times 4096}$ are important for capturing the actions and temporal interactions, and are usually given by the output of fc6 layer in the C3D network \cite{Tran2015C3D} pre-trained on Sports-1M dataset \cite{Karpathy2014sports-1M}, where the dimension of extracted features is 4096.
For the audio features, which are helpful for distinguishing events, the Bag-of-Audio-Words (BoAW) \cite{Pancoast2014boaw}, Fisher Vector \cite{Jorge2013FisherVector} and VGGish \cite{Hershey2017VGGish} are introduced to extract the 256-D, 260-D and 128-D audio features, respectively. We use the concatenation of the three extracted audio features as the final audio features $V_\text{A} \in \R^{8 \times 644}$ in our implementation. 
For the textual features $V_\text{T}$, we extract textual concepts using the concept extractor proposed by \cite{Pan2017Attributes}.
At last, the dimension of the extracted textual, image, motion and audio features, will all be projected to $d_h = 512$.

\subsubsection{Training Details}
\label{sec:training_details}
To train the UVC-VI, we first use a video captioning paired dataset to train the video-to-pivot captioning model as well as the MCE, and then employ a machine translation dataset to train the pivot-to-target translation model; Next, we freeze the parameters of the captioning and translation models, and use the proposed pseudo supervised training and adversarial training (see Eq.~(\ref{eq:full_loss})) to further train the VIM.
At the inference phase, we employ Eq.~(\ref{eq:uvc-vi}) to generate the target sentence in an end-to-end inference manner.
To implement the pivot-to-target machine translation model and the pivot language decoder of the captioning model, we directly employ the Transformer-BASE model \cite{Vaswani2017transformer}, which achieves great success in machine translation.
Accordingly, the model size $d_h$ is set to 512.
For the MCE, we adopt the multi-head attention \cite{Vaswani2017transformer}, where the number of heads is set to 8. 
According to the performance on the validation set, we set the $\alpha = 0.1$ and $\beta = 10$. 

For the discriminators, following \cite{Yu2017seqgan}, the embeddings will be first projected into 1-D, making the dimension to be 1.
Specifically, in the Adversarial Training stage, we apply the same techniques in \cite{Zhu2017unpairedI2I} to stabilize the adversarial training of VIM and discriminator $\text{D}$, e.g., we train the $\text{VIM}$ by minimizing:
$\E_{V \sim D_1}\left[(\text{D}(\text{VIM} (V))-1)^2\right]$
and the $\text{D}$ by minimizing:
$\E_{T_2 \sim D_2}\left[(\text{D}(T_2)-1)^2\right]$ + 
$\E_{V \sim D_1} \left[\text{D}(\text{VIM}(V))^2\right]$.
Moreover, to further stabilize the Adversarial Training procedure, we first use our proposed Pseudo Supervised Training to pre-train our approach for 50 epochs to initialize proper parameter weights.
Then, we incorporate the Adversarial Training \cite{Zhu2017unpairedI2I} to further train the model for 100 epochs.
We use Adam \cite{kingma2014adam} for parameter optimization.
The learning rate is set to 2e-4.
Meanwhile, we also use momentum of 0.8 and weight decay of 0.999.

For the pivot-to-target translation model, we adopt the Transformer-BASE model \cite{Vaswani2017transformer,ott2019fairseq}.
Following common practice \cite{wu2019paylessattention,Vaswani2017transformer}, we use the \textit{fairseq} \cite{ott2019fairseq} for our implementation of pivot-to-target machine translation model.

\subsection{Comparing Methods}
\subsubsection{Conventional Supervised Systems}
We first compare our approach with several conventional supervised methods, i.e., AF \cite{Hori2017Attention}, MA-LSTM \cite{Xu2017MA-LSTM}, Two-stream \cite{Gao2020Hierarchical}, PickNet \cite{Chen2018PickNet}, RecNet \cite{Zhang2020Reconstruct}, TDConvED \cite{Chen2019Temporal}, POS-CG \cite{Wang2019Controllable}, STAT \cite{Yan2020STAT}, GRU-EVE \cite{Aafaq2019GRU-EVE}, SAAT \cite{Zheng2020SAAT}, SGN \cite{Ryu2021Semantic} and MGSA \cite{Chen2019MGSA}.
These supervised methods follow the common encoder-decoder architecture, trained on large-scale pairs of video and target caption.
The results of these methods are copied from their papers except POS-CG, whose results are copied from \cite{Zheng2020SAAT}.

\subsubsection{Pipeline Systems}
\label{sec:pipeline_compare}
For fair comparison, we investigate the ablative structure of our UVC-VI, i.e., \textbf{Base Model}. As shown in the left plot of Figure~\ref{fig:model}), we directly assemble a video-to-pivot captioning model and a pivot-to-target translation model together. Base Model shares the same structures with the UVC-VI (see Section~\ref{sec:training_details}), i.e., the pivot-to-target machine translation model and the pivot language decoder of the video-to-pivot captioning model are implemented by Transformer-BASE~\cite{Vaswani2017transformer}.
Meanwhile, the Base Model exploits the image+motion and image+motion+audio features on MSVD and MSR-VTT datasets, respectively.
Besides, to improve the evaluation significantly, we further implement three conventional supervised video captioning models equipped with the Google Translator\footnote{Our preliminary experiments show that using Google Translator could achieve better performance than the Transformer-BASE from \textit{fairseq} used in our approach.} \cite{Wu2016google}, acquiring three pipeline systems: 1) \textbf{MA-LSTM \cite{Xu2017MA-LSTM} + Google Translator \cite{Wu2016google}}, 2) \textbf{SAAT \cite{Zheng2020SAAT} + Google Translator \cite{Wu2016google}} and 3) \textbf{SGN \cite{Ryu2021Semantic} + Google Translator \cite{Wu2016google}}.

\begin{table*}[t]
    \centering
    \footnotesize  
     \caption{Performance of human evaluation for comparing ``UVC-VI'' with ``Base Model'' and ``SGN \cite{Ryu2021Semantic} + Google Translator \cite{Wu2016google}'', under three language application scenarios, i.e., French, German and Chinese, where the video-caption pairs in target language are not available. Different from automatic metrics in Table~\ref{tab:unpaired_auto}, the human evaluation results verify the advantage of UVC-VI from the perspective of user experience and prove the effectiveness of UVC-VI in mitigating the visual irrelevancy and disfluency problems  in  pipeline systems.
     \label{tab:unpaired_human}}
    \begin{tabular}{@{}l l c c c c c c@{}}
        \toprule
        \multirow{3}{*}[-3pt]{Metrics} &\multirow{3}{*}[-3pt]{Languages} &   \multicolumn{6}{c}{Setting: Unpaired Video Captioning} \\ \cmidrule(l){3-5} \cmidrule(l){6-8}
        & & ``Base Model'' & \multirow{2}{*}{Tie (\%)} & ``UVC-VI''  & ``SGN \cite{Ryu2021Semantic} + Google Translator \cite{Wu2016google}''\ssymbol{3} & \multirow{2}{*}{Tie (\%)}  & ``UVC-VI''  \\
         & & wins (\%) & & wins (\%) & wins (\%) & & wins (\%) \\ \midrule[\heavyrulewidth]

        \multirow{3}{*}{Fluency} & French  & 26.0  & 45.5  & \bf  \underline{\underline{28.5}} & 29.5 & 36.5 & \bf \underline{\underline{34.0}} \\
        & German & 31.0 & 32.0 & \bf \underline{\underline{37.0}} & 32.5 & 33.0 & \bf \underline{\underline{34.5}} \\
        & Chinese & 28.0 & 35.5 & \bf \underline{\underline{36.5}} & 32.0 & 29.5 & \bf \underline{\underline{38.5}} \\

        \midrule
        
        \multirow{3}{*}{Visual Relevance} & French & 15.5 & 25.5 & \bf \underline{\underline{59.0}} & 17.5 & 37.5 & \bf \underline{\underline{45.0}}\\
        & German &  24.5 & 26.5 & \bf \underline{\underline{49.0}} & 26.5 & 27.0 & \bf \underline{\underline{46.5}} \\
        & Chinese & 19.5& 39.0  & \bf \underline{\underline{41.5}} & 20.5 & 26.5 & \bf \underline{\underline{53.0}} \\ 
       
        \bottomrule
    \end{tabular}
\end{table*}

\subsection{Automatic Evaluation}
\label{sec:auto}

\subsubsection{Datasets}
To conduct the automatic evaluation, we verify the effectiveness of our approach on the benchmark MSVD \cite{Guadarrama2013MSVD} and MSR-VTT \cite{Xu2016MSR-VTT} video English captioning datasets.
In particular, for the MSR-VTT dataset, it contains 10,000 video clips, and each video is paired with 20 annotated sentences. We use the official splits to report our results \cite{Xu2016MSR-VTT}. Thus, there are 6513, 497 and 2990 video clips in training, validation and test set, respectively. 
For MSVD, it contains 1,970 video clips and roughly 80,000 English sentences. We follow the split settings in \cite{Pei2019MARN}, resulting in 1,200, 100 and 670 videos for the training set, validation set and test set, respectively.
It is worth noting that we focus on the unpaired video captioning, where the video-caption pairs in target language are not available, the video-caption pairs in the training sets of MSVD and MSR-VTT are discarded and are not used in our training stage.

In implementation, to train our UVC-VI, we take Chinese as the pivot language and train the video-to-pivot captioning model on VATEX video Chinese captioning dataset \cite{Wang2019VATEX}, which contains 41,250 video clips, and each video clip contains 10 human-annotated Chinese captions. 
We train the pivot-to-target translation model on the WMT17 Chinese-English dataset\footnote{\url{http://statmt.org/wmt17/}}.
We follow their official settings to pre-process the WMT17 and VATEX datasets.
In our experiment, we further employ the target captions from training sets to train the decoder of the pivot-to-target model by reconstructing the target captions in an auto-encoding pipeline.

\subsubsection{Metrics}
We report the results using the standard automatic evaluation toolkit \cite{chen2015microsoft}, which includes the widely-used metrics CIDEr \cite{vedantam2015cider}, ROUGE-L \cite{lin2004rouge}, METEOR \cite{banerjee2005meteor} and BLEU \cite{papineni2002bleu}. 
CIDEr, which is based on n-gram matching, incorporating the consensus of a reference set for an example.
ROUGE-L is proposed for automatic evaluation of the extracted text summarization. 
METEOR and BLEU are originally designed for machine translation evaluation. 
Among them, the CIDEr is specifically designed for evaluating captioning systems and will be the main considered metric.

\subsubsection{Results} 
As mentioned in Section~\ref{sec:pipeline_compare}, in comparable settings, i.e., unpaired video captioning, to improve the evaluation significantly, we further implement three representative video captioning models equipped with the Google Translator \cite{Wu2016google}, resulting in three pipeline systems: ``MA-LSTM + Google Translator'', ``SAAT + Google Translator'' and ``SGN + Google Translator''.
The performance of the three implemented pipeline systems and our baseline method on the test set of MSVD and MSR-VTT datasets are shown in Table~\ref{tab:unpaired_auto}.
As we can see, both of them are significantly lower than our proposed UVC-VI across most metrics, which demonstrates the effectiveness of our approach.
The reason is that our UVC-VI can simplify the entire workflow and enable visual information to directly reach the translation model.
As a result, our UVC-VI can address the visual irrelevancy and disfluency errors in the pipeline systems.

More encouragingly, our UVC-VI (I+M) and UVC-VI (I+M+A) can surpass several supervised video captioning models, which consist of some recently supervised models, i.e., Two-stream$_\text{(TPAMI2020)}$ \cite{Gao2020Hierarchical}, RecNet$_\text{(TPAMI2020)}$ \cite{Zhang2020Reconstruct}, POS-CG$_\text{(ICCV2019)}$ \cite{Wang2019Controllable}, STAT$_\text{(TMM2020)}$ \cite{Yan2020STAT},  GRU-EVE$_\text{(CVPR2019)}$ \cite{Aafaq2019GRU-EVE} and SAAT$_\text{(CVPR2020)}$ \cite{Zheng2020SAAT}.
It is worth noting that the POS-CG$_\text{(ICCV2019)}$ \cite{Wang2019Controllable}, GRU-EVE$_\text{(CVPR2019)}$ \cite{Aafaq2019GRU-EVE} and SAAT$_\text{(CVPR2020)}$ \cite{Zheng2020SAAT} adopt the same features, i.e., Inception-ResNet-V2 \cite{Szegedy2017Inception-v4} for image features and C3D \cite{Tran2015C3D} for motion features, with our UVC-VI (I+M), the superiority of our approach using same features further proves our arguments and corroborates the effectiveness of our approach, which can relax the reliance on the paired dataset for video captioning.

The experimental results show that our approach is able to generate fluent and desirable video captions without the training on the pairs of video and target caption, which could have the potential to promote the application of video captioning for various low-resource language applications.

\subsection{Human Evaluation}
\label{sec:human}
\subsubsection{Datasets}
In this section, we further conduct the experiments on three language application scenarios, i.e., French, German and Chinese.
In implementation, we take English as the pivot language and train the video-to-pivot captioning model on the MSR-VTT video English captioning dataset \cite{Xu2016MSR-VTT}.
For different language application scenarios, we train the pivot-to-target translation model on WMT translation datasets. Specifically, we adopt the WMT14 English-French dataset for French, WMT14 English-German dataset for German\footnote{\url{http://statmt.org/wmt14/}}, and WMT17 English-Chinese dataset for Chinese\footnote{\url{http://statmt.org/wmt17/}}.
Specifically, WMT14 English-German, WMT14 English-French and WMT17 English-Chinese consist of 4.5M, 36M and 24M sentence pairs, respectively.

\begin{table}[t]

\centering
\footnotesize
\caption{Quantitative Analysis of our approach which is performed on the test set of MSR-VTT video captioning dataset \cite{Xu2016MSR-VTT} under the unpaired setting. The \textbf{\underline{\underline{Bold}}} numbers denote the best results across all methods.
\label{tab:quantitative_analysis}}
\setlength{\tabcolsep}{2pt}
\begin{tabular}{@{}l c c c c c c c@{}}
\toprule
{Methods}  
& {${L}_{\text{pseudo}}$  }
& {${L}_{\text{adv}}$}   
& {${L}'_{\text{adv}}$} 
& {${L}_{\text{cyc}}$} 
 & BLEU-4 & METEOR & CIDEr \\ \midrule [\heavyrulewidth]

\multicolumn{8}{c}{Setting: Unpaired Video Captioning} \\ \midrule

Base Model  & & & & & 34.7 & 25.1 & 36.3 \\ \midrule

\ w/ VIM &$\surd$  & & &  & 36.8 & 26.6 & 40.7 \\

\ w/ VIM &$\surd$ &$\surd$ & &  & 38.2 & 27.2 & 42.0 \\

\ w/ VIM &$\surd$ &$\surd$ &$\surd$ &$\surd$ & 38.5 & 27.4 & 42.3 \\ 
\midrule

\begin{tabular}[c]{@{}c@{}} \ w/ VIM + MCE \\ (UVC-VI) \end{tabular} &$\surd$ &$\surd$ &$\surd$ &$\surd$ & \bf \underline{\underline{38.9}} & \bf \underline{\underline{27.8}} & \bf \underline{\underline{44.5}}   \\
\bottomrule
\end{tabular}
\end{table}
\begin{figure}[t]

\centering
\includegraphics[width=1\linewidth]{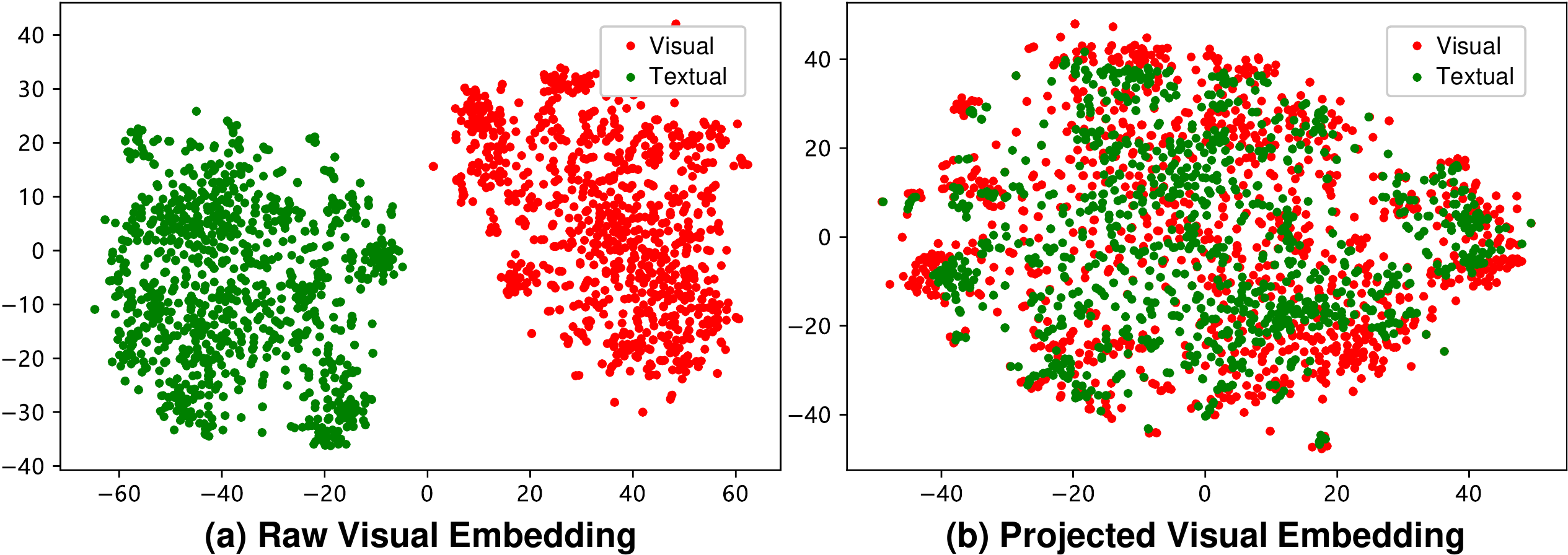}

\caption{Visualization of (a) the raw visual embedding $V$ and (b) projected visual embedding $\text{VIM}(V)$ in 2D space by t-SNE \cite{hinton2008t-sne} (see Eq.~(\ref{eqn:vim})). For comparison, we show the textual embedding $T$ in the target language domain. We plot the scatter diagrams for 1,000 samples. As we can see, our VIM can align the latent space of $V$ and $T$ effectively, i.e., successfully project the visual embedding of source video domain into the target language domain.}
\label{fig:injection_vis}
\end{figure}

\subsubsection{Metrics}
Due to the lack of video-caption pairs in the target language domain, to prove our arguments and evaluate the effectiveness of our approach in mitigating the \textit{visual irrelevancy} and \textit{disfluency} problems on the target language, for each target language, we randomly selected 200 videos from the MSR-VTT dataset, and recruit 10 annotators with sufficient language skills (5 proficient in French, 4 in German and 1 in Chinese), to conduct the human evaluation. Each annotator is required to evaluate the perceptual quality, including the \textit{visual relevance} and \textit{fluency}, of the generated captions in the target language and compare the performance of UVC-VI with the baselines.
They have no idea which model the captions are produced by. The results are calculated by the picking percentages (\%).

\subsubsection{Results}
To obtain the results of human evaluation, we select the ``Base Model'' and the ``SGN \cite{Ryu2021Semantic} + Google Translator \cite{Wu2016google}''.
The results in Table~\ref{tab:unpaired_human} show that the captions generated by our UVC-VI are more favored by annotators than the two selected baseline approaches on all metrics across all language application scenarios. It verifies our arguments and proves the effectiveness of our approach in mitigating the visual irrelevancy and disfluency problems in pipeline systems.
In particular, our approach achieves better improvements in terms of the \textit{visual relevance} than the \textit{fluency}, which indicates that the visual relevance problem is the more critical in the pipeline systems. The reason is that, in the pipeline systems, the visual information cannot reach the translation model at all, which further proves the effectiveness of our approach.

From the results of automatic and human evaluations, we can see that our UVC-VI can well-perform the unpaired video captioning, regardless of the downstream language application scenarios.
Especially, our approach can outperform several supervised video captioning models.
As a result, since our model dose not rely on any video-target paired data, our work could improve the practicality of video captioning in real-world applications, such as video retrieval \cite{Yu2017Retrieval} and visually impaired people aiding \cite{Voykinska2016helpsee}, especially for non-English languages applications.

\section{Analysis}
\label{sec:quantitative}
In this section, to better understand our methods, we conduct the analysis to investigate the contribution of each component in our proposed model, i.e., Visual Injection Module (VIM) and Multimodal Collaborative Encoder (MCE).

\subsection{Effect of Visual Injection Module (VIM)}
In this section, we analyze the effect of VIM, including the proposed pseudo supervised training and adversarial training.
From Table~\ref{tab:quantitative_analysis}, we can see that the ${L}_{\text{pseudo}}$ provides a solid basis for VIM to inject the visual information into language domain. 
Since adversarial training can further make use of the pivot-target paired dataset, the introduction of ${L}_{\text{adv}}$ further improves the performance.
By introducing the cycle strategy (i.e., ${L}'_{\text{adv}}$ + ${L}_{\text{cyc}}$), the performance improves as well. The reason may be that the cycle strategy enforces ${L}_{\text{adv}}$ and ${L}'_{\text{adv}}$ to collaborate with each other \cite{Zhu2017unpairedI2I}.

To better understand our approach, we further report the breakdown of SPICE F-scores \cite{anderson2016spice} in Table~\ref{tab:spice}.
As we can see, comparing ``w/ VIM'' and ``Base Model'' in SPICE F-score, the former significantly achieves better \textit{Attributes} and \textit{Color} scores, this is due to the enriched visual details contained in the visual embedding, which are injected into target language domain by our VIM.

To further verify the VIM indeed aligns the source visual and target language domains, following \cite{gu2019unpaired} and \cite{Laina2019Towards}, we adopt the t-SNE \cite{hinton2008t-sne} to visualize the features in Figure~\ref{fig:injection_vis}, which shows that the VIM actually injects the visual information into the latent space of target language domain, resulting in a blended distribution.

\begin{table}[t]

\centering
\footnotesize
        \caption{We further report the SPICE F-scores \cite{anderson2016spice} on the test set of MSR-VTT dataset for a better understanding of the differences of the target captions generated by different methods.}
        \begin{center}
        \begin{tabular}{@{}l c c c c c@{}}
        \toprule
        \multirow{2}{*}[-3pt]{Methods}  
        & \multicolumn{5}{c}{SPICE}  \\ \cmidrule(lr){2-6}  & All & Objects & Attributes & Relations & Color  \\ \midrule  [\heavyrulewidth] 
        \multicolumn{6}{c}{Setting: Unpaired Video Captioning}
        \\ \midrule
        
        Base Model & 6.1 & 13.0 & 1.9 & 2.3 & 0.1 \\  \midrule
        \ w/ VIM  & 6.5 & 13.3 & 2.4 & \bf \underline{\underline{2.4}} & 0.6  \\
        
        \midrule
        \begin{tabular}[c]{@{}c@{}} \ w/ VIM + MCE\\ (UVC-VI) \end{tabular} & \bf \underline{\underline{6.7}}  & \bf \underline{\underline{13.9}} & \bf \underline{\underline{2.5}} & \bf \underline{\underline{2.4}} & \bf \underline{\underline{0.7}}\\

        \bottomrule
        \end{tabular}
        \end{center}
        \label{tab:spice}
\end{table}

\begin{table*}[t]

\centering
\footnotesize
\caption{Analysis of the Multimodal Collaborative Encoder (MCE). As a pluggable video encoder, MCE can be easily integrated into existing video captioning models. Therefore, we further equip our MCE with two representative models, i.e., Attention-based LSTM model (Attention-LSTM) \cite{anderson2018bottom} and Transformer model \cite{Vaswani2017transformer}. We perform the analysis on the MSVD and MSR-VTT datasets under the paired setting, i.e., conventional supervised video captioning. 
Symbol \ssymbol{3} is defined similarly.
The \textbf{\underline{\underline{Bold}}} numbers denote the best results across all methods. As we can see, our MCE can successfully boost the baselines, with the most improvement up to relatively 7\% and 11\% for MSVD and MSR-VTT in terms of CIDEr, which is specifically designed for evaluating captioning systems, respectively. More encouragingly, our Transformer w/ MCE model surpasses the previous state-of-the-art models on the MSVD and MSR-VTT, with relatively 4\% and 7\% margins and in terms of the CIDEr, respectively.}
\setlength{\tabcolsep}{5pt}

\begin{center}
\begin{tabular}{@{}l l c c c c l c c c c@{}}
\toprule
\multirow{2}{*}[-3pt]{Methods}  
& \multicolumn{5}{c}{Dataset: MSVD \cite{Guadarrama2013MSVD}} & \multicolumn{5}{c}{Dataset: MSR-VTT \cite{Xu2016MSR-VTT}}  \\ \cmidrule(lr){2-6} \cmidrule(lr){7-11}  & Features & BLEU-4 & METEOR & ROUGE-L & CIDEr & Features & BLEU-4 & METEOR & ROUGE-L & CIDEr \\ \midrule  [\heavyrulewidth] \multicolumn{11}{c}{Setting: Conventional Supervised Video Captioning} \\ \midrule

Two-stream$_\text{(TPAMI2020)}$ \cite{Gao2020Hierarchical} & I+M &54.3 &33.5 & - & 72.8 & I+M & 39.7 & 27.0 & - & 42.1 \\
RecNet$_\text{(TPAMI2020)}$ \cite{Zhang2020Reconstruct} & I & 52.3 & 34.1 & 69.8 & 80.3 & I & 39.1 & 26.6 & 59.3 & 42.7 \\
STAT$_\text{(TMM2020)}$ \cite{Yan2020STAT} & I+M & 52.0 & 33.3 & - & 73.8 & I+M & 39.3 & 27.1 & - & 43.8 \\
GRU-EVE$_\text{(CVPR2019)}$ \cite{Aafaq2019GRU-EVE} & I+M &  47.9 & 35.0 & 71.5 & 78.1 & I+M & 38.3 &28.4 & 60.7 & 48.1 \\ 
SibNet$_\text{(TPAMI2020)}$ \cite{Liu2020SibNet} & I & \bf\underline{\underline{55.7}} & 35.5 &72.6 & 88.8 & I & 41.2 &27.8 &60.8 &48.6\\
SAAT$_\text{(CVPR2020)}$ \cite{Zheng2020SAAT} & I+M & 46.5 & 33.5 & 69.4 & 81.0 & I+M & 39.9 & 27.7 & 61.2 & 51.0 \\ 
SGN$_\text{(AAAI2021)}$ \cite{Ryu2021Semantic} & I+M &52.8 & 35.5 & 72.9 & 94.3 & I+M & 40.8 & 28.3 & 60.8 & 49.5 \\
MGSA$_\text{(AAAI2019)}$ \cite{Chen2019MGSA} & I+M & 53.4  & 35.0 &- & 86.7  & I+M+A & 45.4 & 28.6 & - & 50.1 \\  

\midrule
[\heavyrulewidth]

Attention-LSTM \cite{anderson2018bottom}\ssymbol{3} & I+M & 51.6 & 35.4 & 71.8 & 88.5 & I+M+A &  44.4 & 29.7 & 62.1 & 48.9 \\
\ w/ MCE & I+M & 52.8  &   35.7  &  72.6  &  93.4  & I+M+A & 44.9&  30.0 & 63.0 &  52.6  \\ \midrule

Transformer \cite{Vaswani2017transformer} \ssymbol{3} & I+M & 49.8 & 35.1 & 72.2 & 91.2 & I+M+A & 43.2 & 28.5 & 61.8 & 49.3   \\ 
\ w/ MCE  & I+M & 55.1 & \bf\underline{\underline{36.2}}&  \bf\underline{\underline{73.3}}&   \bf\underline{\underline{97.7}}& I+M+A & \bf\underline{\underline{45.9}}&  \bf\underline{\underline{30.7}}& \bf\underline{\underline{64.3}}& \bf\underline{\underline{54.6}}\\

\bottomrule
\end{tabular}
\end{center}
\label{tab:MCE}

\end{table*}

\subsection{Effect of Multimodal Collaborative Encoder (MCE)}
\label{sec:Analysis_MCE}
Table~\ref{tab:quantitative_analysis} shows that the MCE can further promote the performance, which verify our arguments and prove the MCE can smooth the cross-modality projection of VIM.
To better understand our MCE, we also report the breakdown of SPICE F-scores \cite{anderson2016spice} in Table~\ref{tab:spice}. It shows that the MCE can significantly improve the \textit{Object} score. It proves that textual-enriched visual embedding produced by MCE can efficiently integrate multimodal features according to each \text{object} in textual concepts.

As a video encoder, MCE can be easily integrated into existing supervised video captioning models. Therefore, we further equip our MCE with two representative models, i.e., Attention-based LSTM model (Attention-LSTM) \cite{anderson2018bottom} and Transformer model \cite{Vaswani2017transformer}. We evaluate the performance of these models on the MSVD and MSR-VTT video captioning datasets, under the paired setting, i.e., conventional supervised video captioning. 
Table~\ref{tab:MCE} shows that MCE can successfully boost the baselines, with the most significant improvement up to relatively {7\%} and {11\%} for MSVD and MSR-VTT in terms of CIDEr, respectively.
Moreover, our approach surpasses the previous state-of-the-art models on the MSVD and MSR-VTT datasets, with relatively {4\%} and {7\%} margins on the CIDEr scores, respectively.
The improvements demonstrate the effectiveness and generalization ability of our proposed MCE.

\subsection{Examples}
\begin{figure*}[t]

\centering
\includegraphics[width=0.6\linewidth,trim={0 108 0 0},clip]{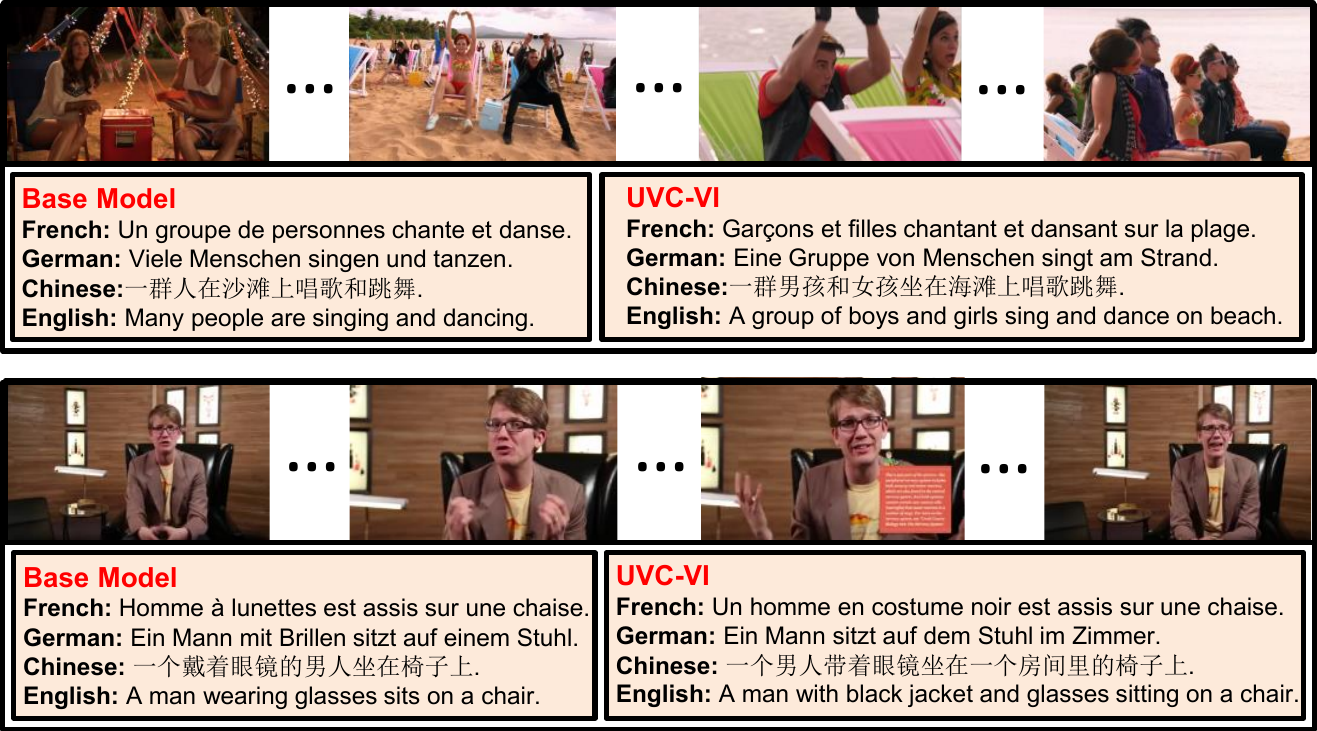}
\caption{The examples of video captions generated by the Base Model and UVC-VI for different languages, i.e., French, German, Chinese and English, under the unpaired setting, where the video-caption pairs in target language are not available. As we can see, without the training on the pairs of video and target caption, our approach can generate fluent and desirable video captions for different languages.}
\label{fig:example}
\end{figure*}

In this section, we provide some examples under different language application scenarios, i.e., English, French, German and Chinese, to show the advantages of UVC-VI from the perspective of user experience.
As mentioned above, in case of English as the target language, we take Chinese as pivot language; for the other three languages, we take English as pivot language.
As shown in Figure~\ref{fig:example}, the generated captions by UVC-VI are more fluent than the pipeline system (i.e., Base Model). Moreover, the proposed UVC-VI includes more objects, such as \textit{beach} and \textit{jacket} in the $1^{\rm st}$ and $2^{\rm nd}$ examples, respectively, and is good at portraying the attribute, e.g., \textit{black} in the $2^{\rm nd}$ example, and describing the details, e.g., \textit{boys} and \textit{girls} in the $1^{\rm st}$ example.
The reason is that the introduction of our Visual Injection Module (VIM) in UVC-VI can enable visual information to directly reach the target language decoder which lowers the chances of error occurrence and amplification.

\subsection{Extension to Unpaired Image Captioning}

Considering the fact that the unpaired image captioning has been explored in the literature \cite{gu2018unpaired,gu2019unpaired,Feng2019Unsupervised_ic,Laina2019Towards,Yang2020USGAE,liu2019unpaired} (see Section~\ref{sec:UIC}), we further extend our UVC-VI to the unpaired image captioning to verify the effectiveness of our approach.
Specifically, following the existing approaches, we also perform the experiments on the MSCOCO image English captioning dataset \cite{chen2015microsoft}.

\subsubsection{Metrics and Datasets}
Following common practice \cite{gu2019unpaired,Laina2019Towards,Yang2020USGAE}, we test the performance of image captioning model with captioning evaluation toolkit \cite{chen2015microsoft}, which reports the widely-used automatic evaluation metrics CIDEr \cite{vedantam2015cider}, ROUGE-L \cite{lin2004rouge}, METEOR \cite{banerjee2005meteor} and BLEU \cite{papineni2002bleu}. 
For fair comparisons, we conduct the evaluation on the popular MSCOCO image English captioning dataset \cite{chen2015microsoft}.
Specifically, we use the publicly-available splits in \cite{karpathy2014deep} for evaluation. There are 5,000 images each in validation set and test set.
During training, following \cite{gu2018unpaired}, we take Chinese as the pivot language, learn the image-to-pivot captioning model on ICC image Chinese captioning dataset \cite{Wu2019AIChallenger} and train the pivot-to-target translation model on the WMT17 Chinese-English dataset.

\subsubsection{Settings} 
To extend our UVC-VI to the unpaired image captioning task, we keep the inner structure of the UVC-VI untouched and directly remove the motion and audio features which are only available in videos, i.e., we use the set ${V} = [{V}_\text{T}; {V}_\text{I}]$ to denote the visual embedding of image.

For fair comparisons, for the image features ${{V}}_\text{I}$, we use the RCNN-based image features provided by \cite{anderson2018bottom}.
The dimension of the extracted RCNN-based image features will all be projected to 512, which is equal to the dimension of our captioning model $d_h = 512$.
For the textual features ${{V}}_\text{T}$, we extract textual concepts using the textual concept extractor proposed by \cite{fang2015captions}.
The embedding vectors (512-D) of these extracted textual concepts are then taken as the ${{V}}_\text{T}$.

\begin{table}[t]
    \centering
    \footnotesize  
    \caption{Performance of our proposed UVC-VI on the test split of MSCOCO image captioning dataset \cite{chen2015microsoft} under the unpaired setting, i.e., unpaired image captioning, where the image-caption pairs in target language are not available. The \textbf{\underline{\underline{Bold}}} numbers denote the best results across all approaches. As we can see, our UVC-VI achieves the best  performance among the existing approaches for unpaired image captioning across all metrics.
    \label{tab:extend_ic}}
    \begin{center}
    \setlength{\tabcolsep}{1.3pt}
    \begin{tabular}{@{}l c c c c@{}}
        \toprule
        \multirow{2}{*}[-3pt]{Methods} &  \multicolumn{4}{c}{Dataset: MSCOCO \cite{chen2015microsoft}}  \\ \cmidrule(lr){2-5}& BLEU-4 & METEOR & ROUGE-L & CIDEr \\ \midrule [\heavyrulewidth]
        \multicolumn{5}{c}{Setting: Unpaired Image Captioning} \\ \midrule
        Language Pivoting$_\text{(ECCV2018)}$ \cite{gu2018unpaired}&  5.4  &  13.2 & -  &  17.7  \\
        Adv.+Recon.$_\text{(CVPR2019)}$ \cite{Feng2019Unsupervised_ic} & 18.6 & 17.9 & 43.1 & 54.9  \\
        Shared Embeddings$_\text{(ICCV2019)}$ \cite{Laina2019Towards} & 19.3 & 20.2 & 45.0 & 61.8  \\ 
        Graph-Align$_\text{(ICCV2019)}$ \cite{gu2019unpaired} & 21.5 & 20.9 & 47.2 & 69.5  \\
        USGAE$_\text{(TPAMI2020)}$ \cite{Yang2020USGAE} & 17.1 & 19.1 &43.8 &55.1 \\\midrule 
        Base Model   & 16.7 & 17.8 & 40.8 & 51.7 \\
        \ w/ VIM & 20.5 & 19.9 & 45.7 & 64.1  \\
        \ w/ VIM + MCE (UVC-VI)  & \bf \underline{\underline{22.0}} & \bf \underline{\underline{21.4}} & \bf \underline{\underline{47.6}} & \bf \underline{\underline{72.3}}\\ 
        \bottomrule
    \end{tabular}
    \end{center}
\end{table}
\begin{figure}[t]
\includegraphics[width=1\linewidth]{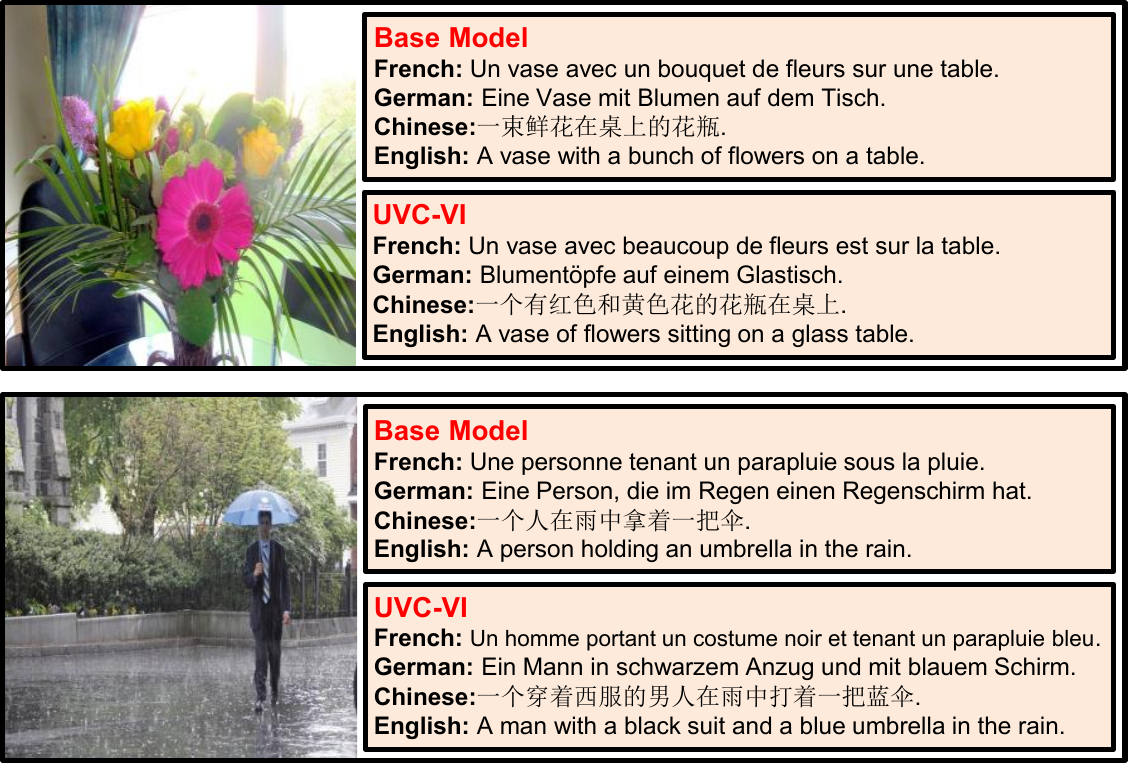}

\caption{The examples of image captions generated by the Base Model and our UVC-VI for different languages, i.e., French, German, Chinese and English, under the unpaired setting. As we can see, our UVC-VI can be extended to unpaired image captioning, where the image-caption pairs in target language are not available, to generate desirable and fluent image captions.}
\label{fig:ic_example}
\end{figure}

\subsubsection{Implementation Details}
Since the image captioning only accept the image features ${{V}}_\text{T}$ and textual concepts ${{V}}_\text{T}$, we should modify the Multimodal Collaborative Encoder (MCE), which is original designed for video captioning, to extend to the image captioning, while the rest of model structure and implementation details are the same as as the UVC-VI used for unpaired video captioning.
In implementations, given the multimodal features ${V} = [{V}_\text{T}; {V}_\text{I}]$, the Multimodal Collaborative Encoder (MCE) for image captioning is defined as:
\begin{equation}
V' = \text{MCE}(V) =  {\gamma} \odot \text{Attention}\left({{V}}_\text{T}, {{V}}_\text{I}\right) ,
\end{equation}
where ${V}'$ denotes the text-enriched visual embedding and the ${\gamma}$ and $\text{Attention}\left(\cdot,\cdot\right)$ are defined as follows:
\begin{equation}
\begin{aligned}
{\gamma} = &  \sigma\left(\left[{{V}}_\text{T}; \text{Attention}\left({{V}}_\text{T}, {{V}}_\text{I}\right)\right]W_{\gamma}\right) , \\
\text{Attention}&(x, y) = \text{softmax}\left(x W_\text{q}\left(y W_\text{k}\right)^{\top}\right) y W_\text{v} .
\end{aligned}
\end{equation}
where $\odot$ and $\sigma$ denote the element-wise multiplication and the sigmoid activation, respectively; $[\cdot; \cdot]$ denotes concatenation operation.
$W_{\gamma} \in \R^{2d_h \times d_h}$, $W_\text{q} \in \R^{d_h \times d_h}$, $W_\text{k} \in \R^{d_h \times d_h}$ and $W_\text{v} \in \R^{d_h \times d_h}$ are learnable parameters.

\subsubsection{Results}
For the existing models for unpaired image captioning \cite{gu2018unpaired,Yang2020USGAE}, where the image-caption pairs in target language are not available, we directly report the results from the original papers.
The quantitative results are illustrated in the Table~\ref{tab:extend_ic}, showing that both the MCE and VIM can boost the baseline.
Furthermore, the UVC-VI achieves the best performance among the existing methods across all metrics.
It further proves the effectiveness of our approach for unpaired visual captioning task.

We further show the image captions generated by the method Base Model and the method UVC-VI in Figure~\ref{fig:ic_example} to intuitively analyze the differences of the methods. 
As we can see, the proposed UVC-VI includes more objects, such as ``\textit{black suit}'' in the $2^{\rm nd}$ example, and is good at portraying the attribute and color, such as ``\textit{glass} table'' and ``\textit{blue} umbrella'' $1^{\rm st}$ and $2^{\rm nd}$ examples, respectively.
Besides, the generated captions by UVC-VI are more fluent, because the introduction of our Visual Injection Module in UVC-VI enables visual information to directly reach the target language decoder and lowers the chances of error generation and amplification.
These examples further prove our arguments and the effectiveness of our approach.

\section{Conclusions}

In this paper, we make the first attempt to conduct unpaired video captioning under various low-resource language application scenarios, e.g., French, German and Chinese, in which the video-caption pairs are not available.
To this end, we present the Unpaired Video Captioning with Visual Injection system (UVC-VI). Different from the Pipeline System, which first utilizes video-to-pivot captioning model to generate captions in pivot language and then utilizes pivot-to-target translation model to translate the pivot captions to the target language, 1) UVC-VI can inject the source visual information into the target language domain by aligning the source visual and target language domains; 
2) UVC-VI directly connects the encoder of the video-to-pivot captioning model and the decoder of the pivot-to-target translation model, the pivot caption is no longer generated, which allows end-to-end inference.
The experiments on two benchmark datasets prove the effectiveness of UVC-VI, which outperforms both pipeline systems and several supervised systems.
More encouragingly, the proposed MCE in UVC-VI can be incorporated into existing video captioning models to boost their performances, up to 4\% and 7\% relative margin in terms of CIDEr scores with current state-of-the-art models on the MSVD and MSR-VTT datasets, respectively.
Moreover, we extend our approach to the unpaired image captioning. The experiments show that UVC-VI obtains positive experimental results.

It can be interesting to apply UVC-VI to improve other pipeline systems. The tasks include broad natural language generation tasks such as machine listening comprehension and spoken question answering, as well as those in computer vision such as medical image classification.

\section*{Acknowledgments}
This paper was partially supported by National Natural Science Foundation of China (NSFC 62176008). Special acknowledgements are given to AOTO-PKUSZ Joint Research Center for its support. More importantly, we would like to sincerely thank all the anonymous reviewers and editors for their constructive comments and suggestions that substantially improved this paper. Xu Sun, Yuexian Zou and Xian Wu are the corresponding authors of this paper.

\ifCLASSOPTIONcaptionsoff
  \newpage
\fi

\bibliographystyle{IEEEtran}
\bibliography{tpami}

%
%
\begin{IEEEbiography}[{\includegraphics[width=1in,height=1.25in,clip,keepaspectratio]{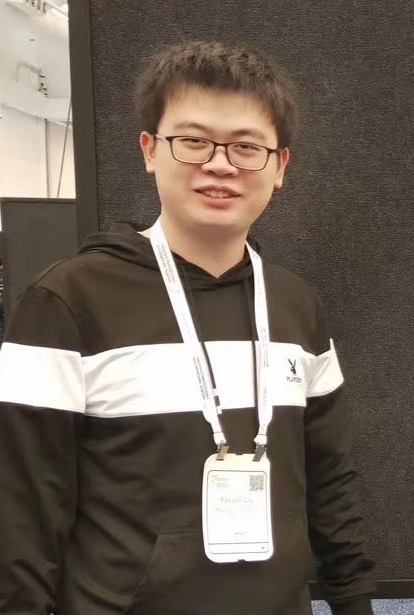}}]{Fenglin Liu} 
is a PhD student at the University of Oxford. His research interests include Natural Language Processing (NLP), especially vision-and-language, Machine Learning, and their applications to clinical, i.e., Clinical NLP.
He has published papers at top-tier journals and conferences, e.g., TPAMI, NeurIPS, CVPR, ACL, EMNLP, NAACL. He has served as a senior program committee member for IJCAI and was awarded as the Distinguished/Outstanding Reviewer of CVPR, AAAI, and IJCAI.

\end{IEEEbiography}


\begin{IEEEbiography}[{\includegraphics[width=1in,height=1.25in,clip,keepaspectratio]{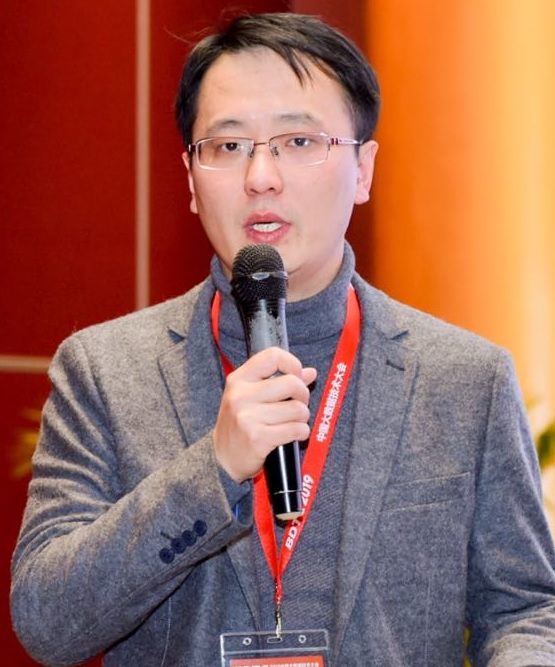}}]{Xian Wu} is now a Principal Researcher in Tencent. Before joining Tencent, he worked as a Senior Scientist Manager and a Staff Researcher in Microsoft and IBM Research. Xian Wu received his PhD degree from Shanghai Jiao Tong University. His research interests includes Medical AI, Natural Language Processing and Multi-Modal modeling. Xian Wu has published papers in CVPR, NeurIPS, ACL, WWW, AAAI, IJCAI etc. He also served as PC member of TKDE, TKDD, TOIS, TIST, CVPR, ICCV, AAAI etc.
\end{IEEEbiography}

\begin{IEEEbiography}[{\includegraphics[width=1in,height=1.25in,clip,keepaspectratio]{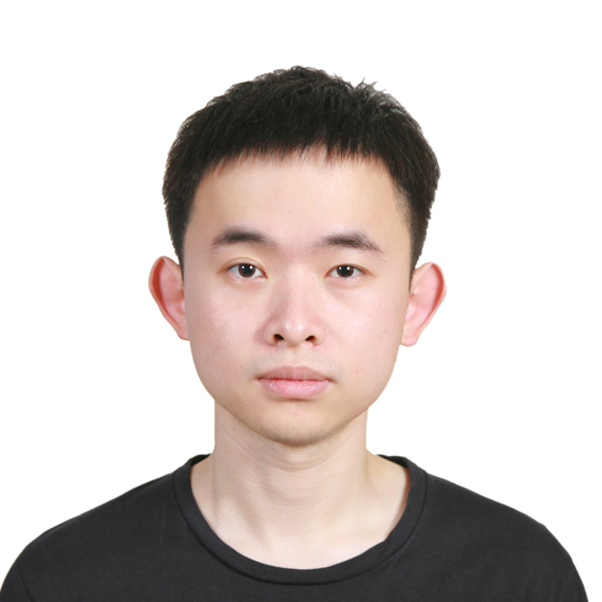}}]{Chenyu You} is a Ph.D. student at Yale University. He received the B.S. and M.Sc. degree from Rensselaer Polytechnic Institute and Stanford University, respectively. His research interests are broadly in the area of machine learning theory and algorithms intersecting the fields of computer \& medical vision, natural language processing, and signal processing. He has been awarded as the Outstanding/Distinguished Reviewer of CVPR, MICCAI, IEEE Transactions on Medical Imaging (TMI), and Medical Physics.
\end{IEEEbiography}


\begin{IEEEbiography}[{\includegraphics[width=1in,height=1.25in,clip,keepaspectratio]{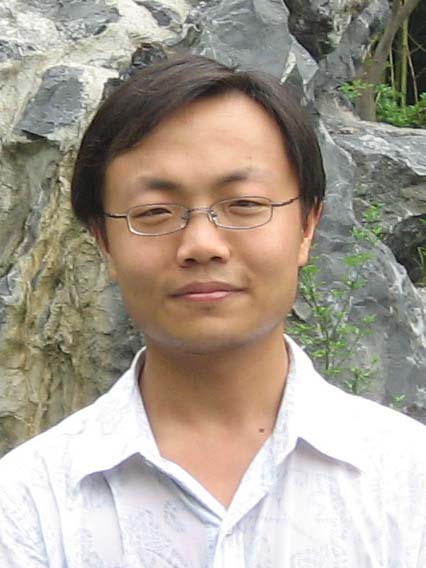}}]{Shen Ge}
Shen Ge is a Senior Researcher in Tencent. His research interests include Natural Language Processing, Deep Learning and Medical AI-based systems. Previously, he has published several works in AAAI, IJCAI, NeurIPS, CVPR, ACMMM, TKDE and KAIS, etc. Prior to Tencent, he worked in DAMO Academy of Alibaba and Microsoft. He received his PhD's degree from the University of Hong Kong, in 2012.
\end{IEEEbiography}

\begin{IEEEbiography}[{\includegraphics[width=1in,height=1.25in,clip,keepaspectratio]{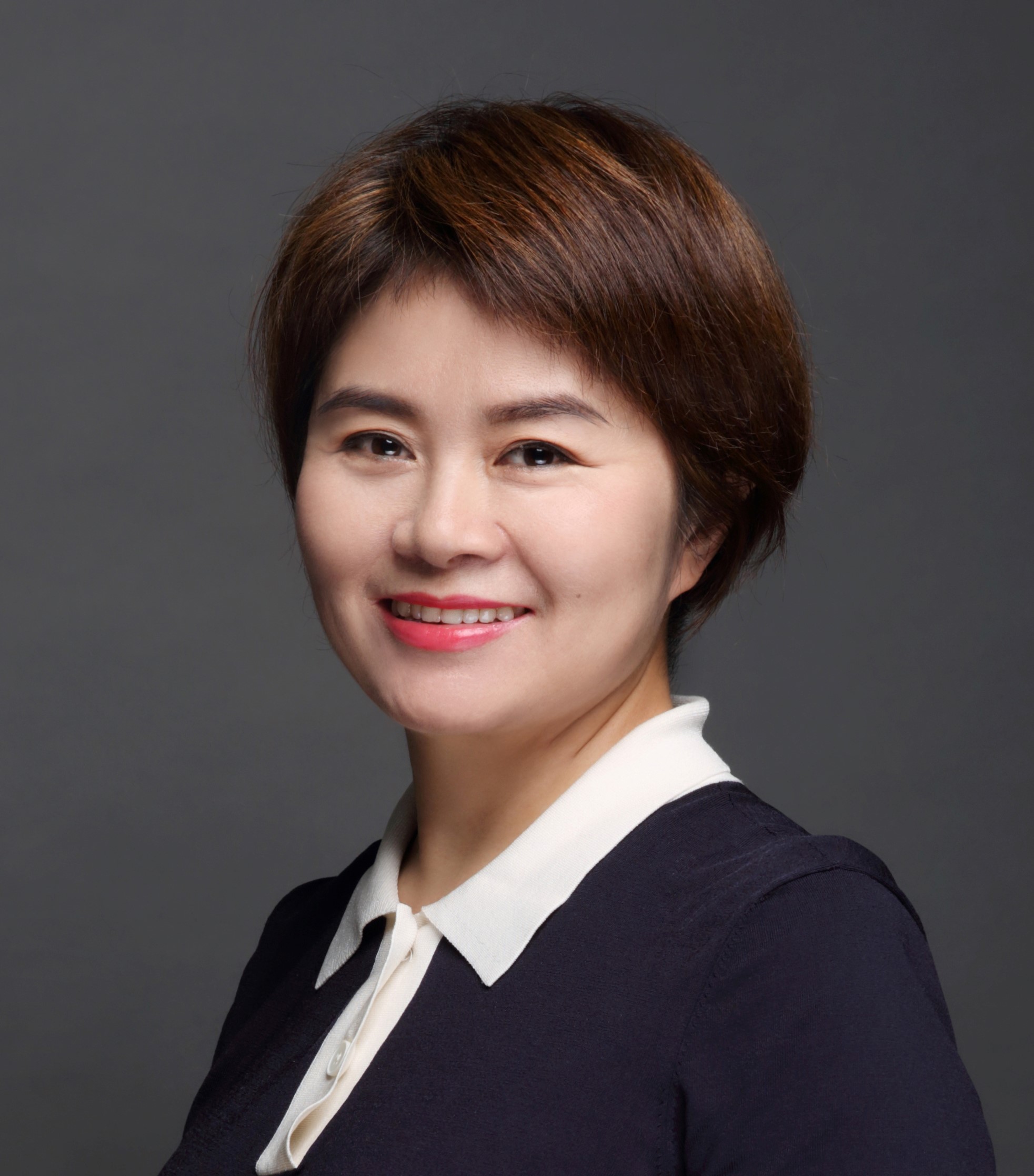}}]{Yuexian Zou}
received her B.Sc. degree from the University of Electronic Science and Technology in 1985 and Ph.D. degree from the University of Hong Kong in 2001, respectively. She is currently a Full Professor with Peking University and the Director of the Advanced Data and Signal Processing Laboratory in Peking University Shenzhen Graduate School. She was a recipient of the award Leading Figure for Science and Technology by Shenzhen Municipal Government in 2009 and now is the adjunct professor in Peng Cheng Laboratory. She also serves as the Deputy Director of Shenzhen Association of Artificial Intelligence (SAAI). Since 2010, she has been actively involved in teaching and research on machine learning and its applications in video and audio analysis. She conducted more than 20 research projects including NSFC and 863 projects. She has published more than 230 academic papers in famous journals and flagship conferences, issued eight invention patents, and two of them have been transferred to a company. She conducts several courses for graduate students, such as machine learning and pattern recognition, digital signal processing, and array signal processing. Her research interests mainly in machine learning for signal processing and scene understanding.
\end{IEEEbiography}


\begin{IEEEbiography}[{\includegraphics[width=1in,height=1.25in,clip,keepaspectratio]{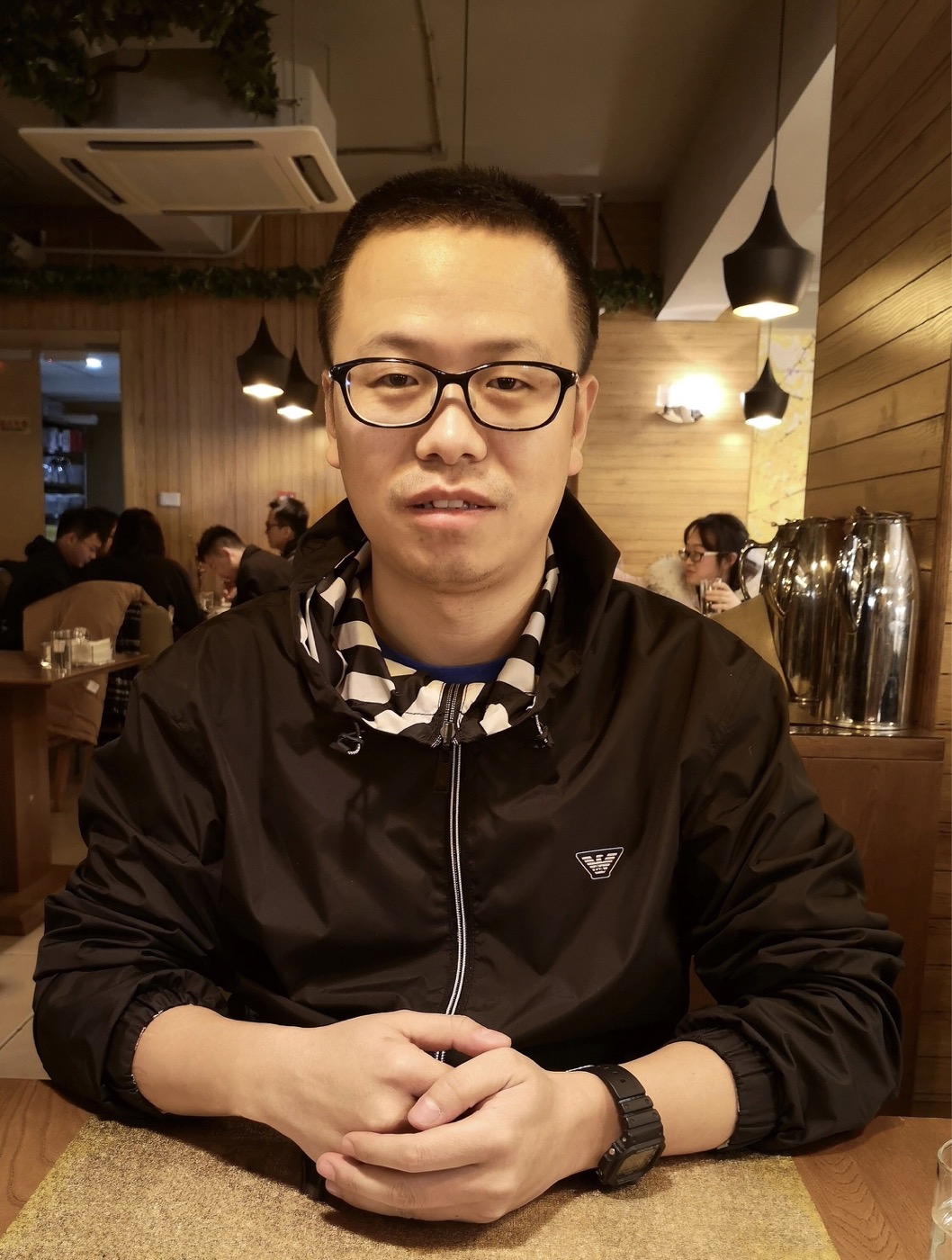}}]{Xu Sun}
is an associate professor in the Department of Computer Science, Peking University. His research focuses on natural language processing and machine learning, especially on structured learning for language, and natural language generation. He has been senior area chair or area chair of ACL, EMNLP, etc. He received COLING Best Paper Award in 2018.
\end{IEEEbiography}





\end{document}